\documentclass[letterpaper]{article} % DO NOT CHANGE THIS

\usepackage{aaai25}  % DO NOT CHANGE THIS
\usepackage{times}  % DO NOT CHANGE THIS
\usepackage{helvet}  % DO NOT CHANGE THIS
\usepackage{courier}  % DO NOT CHANGE THIS
\usepackage[hyphens]{url}  % DO NOT CHANGE THIS
\usepackage{graphicx} % DO NOT CHANGE THIS
\usepackage{natbib}  % DO NOT CHANGE THIS AND DO NOT ADD ANY OPTIONS TO IT
\usepackage{caption} % DO NOT CHANGE THIS AND DO NOT ADD ANY OPTIONS TO IT
\usepackage{algorithm}
\usepackage{algorithmic}
\usepackage{mathtools}
\usepackage{amsmath}
\usepackage{amssymb}

\usepackage{subfigure}
\usepackage{graphicx}
\usepackage{subcaption}
\newcommand{\modelname}{BiMSGC}
\usepackage{newfloat}
\usepackage{listings}
\DeclareCaptionStyle{ruled}{labelfont=normalfont,labelsep=colon,strut=off} % DO NOT CHANGE THIS
\floatstyle{ruled}
\newfloat{listing}{tb}{lst}{}
\floatname{listing}{Listing}
\usepackage[table,xcdraw]{xcolor}
\usepackage{booktabs}
\usepackage{multirow}

\title{Bi-Directional Multi-Scale Graph Dataset Condensation via \\
Information Bottleneck}
\author {
    Xingcheng Fu\textsuperscript{\rm 1, \rm 2, \equalcontrib}\thanks{Corresponding author},
    Yisen Gao\textsuperscript{\rm 3, \equalcontrib},
    Beining Yang\textsuperscript{\rm 5},
    Yuxuan Wu\textsuperscript{\rm 3},
    Haodong Qian\textsuperscript{\rm 1, \rm 2}, \\
    Qingyun Sun\textsuperscript{\rm 4},
    Xianxian Li\textsuperscript{\rm 1, \rm 2}
}
\affiliations {
    \textsuperscript{\rm 1}Guangxi Key Lab of Multi-source Information Mining \& Security, Guangxi Normal University, Guilin, China\\
    \textsuperscript{\rm 2}Key Lab of Education Blockchain and Intelligent Technology, Ministry of Education, Guangxi Normal University, China\\
    \textsuperscript{\rm 3}Institute of Artificial Intelligence, Beihang University, Beijing, China\\
    \textsuperscript{\rm 4}School of Computer Science and Engineering, Beihang University, Beijing, China\\
    \textsuperscript{\rm 5}University of Edinburgh, Edinburgh, UK\\
    \{fuxc, qianhaodong, lixx\}@gxnu.edu.cn \{yisengao, buaawyx, sunqy\}@buaa.edu.cn, B.Yang-32@sms.ed.ac.uk
}

\usepackage{bibentry}
\begin{document}

\maketitle

\begin{abstract}
% dataset condensation has significantly improved model training efficiency, but its application on devices with different computing power brings new requirements for different data sizes. 
% Multi-size dataset condensation has recently shown potential in computer vision to improve training efficiency, however it has two critical challenges with the graph with non-Euclidean structure: node feature subset degradation and subgraph structure degradation. 
% For graph data with non-Euclidean structure, how to condense multiple scale graphs at one time is the core to achieve efficient training in different on-devices scenarios.
% However, it has two critical challenges of multi-scale graph condensation: node feature subset degradation and subgraph structure degradation. 
% To solve the above challenges, 
 % repeated condensation of each scale may lead to significant computational costs.
Dataset condensation has significantly improved model training efficiency, but its application on devices with different computing power brings new requirements for different data sizes. 
Thus, condensing multiple scale graphs simultaneously is the core of achieving efficient training in different on-device scenarios.
Existing efficient works for multi-scale graph dataset condensation mainly perform efficient approximate computation in scale order (large-to-small or small-to-large scales).
However, for non-Euclidean structures of sparse graph data, these two commonly used paradigms for multi-scale graph dataset condensation have serious \textit{scaling down degradation} and \textit{scaling up collapse}  problems of a graph.   
The main bottleneck of the above paradigms is whether the effective information of the original graph is fully preserved when consenting to the primary sub-scale (the first of multiple scales), which determines the condensation effect and consistency of all scales.
In this paper, we proposed a novel GNN-centric \textbf{Bi}-directional \textbf{M}ulti-\textbf{S}cale \textbf{G}raph Dataset \textbf{C}ondensation (\textbf{\modelname}) framework, to explore unifying paradigms by operating on both large-to-small and small-to-large for multi-scale graph condensation. 
% Based on the information bottleneck principle, we estimate an optimal ``\textit{meso-scale}'' to obtain the minimum necessary dense graph preserving the maximum utility information of the original graph, and then we achieve stable and consistent ``\textit{bi-directional}'' condensation learning by graph feature base matching on other scales.
Based on the mutual information theory, we estimate an optimal ``\textit{meso-scale}'' to obtain the minimum necessary dense graph preserving the maximum utility information of the original graph, and then we achieve stable and consistent ``\textit{bi-directional}'' condensation learning by optimizing graph eigenbasis matching with information bottleneck on other scales.
Encouraging empirical results on several datasets demonstrates the significant superiority of the proposed framework in graph condensation at different scales. 

\end{abstract}

\section{Introduction}
% With the extensive growth applications such as social media, e-commerce, and recommendations on mobile and edge devices, the core challenge is how to efficiently train for graph data with exploded scale. 
% With the rapid expansion of applications like social media, e-commerce, and recommendation systems on mobile and edge devices, the core challenge is efficiently training on graph data at an increasingly large scale.
% % 这句需要修改一下
% Recently, dataset condensation has emerged as an important tool  for training on large-scale datasets,  
% demonstrating significant potential in fields such as computer vision and graph learning~\cite{}.
% However, a new challenge for model training is multi-size dataset condensation on devices, the inherent fluctuating computing resources of devices require the flexibility to condense datasets of different sizes at once, and the extra condensation process is unfeasible. 
% For a graph different from the image data, non-Euclidean structures have more topological features at different scales, and these topologies play a key role in the training process for graph learning models. 
% Therefore, the flexibility of different graph scales and the preservation of structural information are the key problems in multi-scale graph compression. 
With the rapid expansion of applications like social media, e-commerce, and recommendation systems on mobile and edge devices, the core challenge is efficiently training on graph data at an increasingly large scale. The multi-size dataset condensation approach is designed to enhance training efficiency by condensing graph datasets into various small yet informative graphs. These condensed graphs can accommodate the diverse computational needs of low-resource, on-device downstream users.
% \begin{figure}[t] % 'htbp' options for positioning
%     \centering
%     \includegraphics[width=0.3\textwidth]{./figs/intro.pdf} % Path to your image
%     \caption{Comparison of two directions}
%     \label{fig:intro}
% \end{figure}

\begin{figure}[t] % 'htbp' options for positioning
    \centering
	\includegraphics[width=0.45\textwidth]{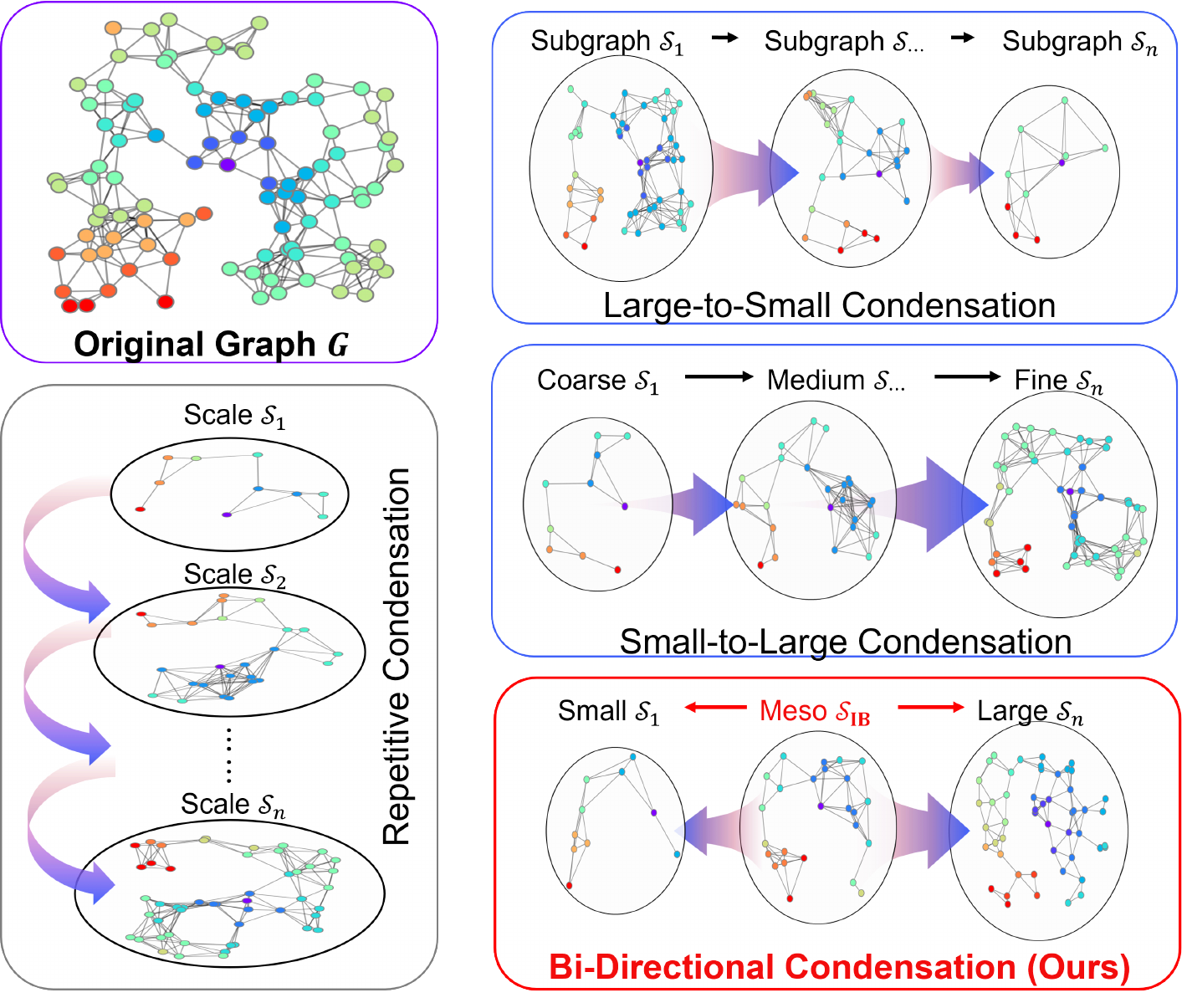}
    \label{fig:intro}
    \caption{Comparison of existing paradigms}
    \label{fig:intro}
\end{figure}

However, dataset condensation typically requires a computationally intensive learning-based process, and the multi-size property exacerbates the challenge in balancing efficiency and effectiveness. Based on the recent studies\cite{he2024multisizedatasetcondensation,fang2024exgc}, there can be summarized three intuitive paradigms (illustrated in the Fig.\ref{fig:intro}): (1) \textit{Re-condensation}, which involves selecting an efficient and effective dataset condensation method to redistribute the data for each scale; (2) \textit{Large-to-Small}, where the original dataset is condensed into multiple size/scales in one process by first creating a relatively large subset and then further condensing it step by step from large to small; and (3) \textit{Small-to-Large}, which starts with the creation of smaller subsets that are then iteratively expanded to larger scales, preserving essential patterns at each stage. These paradigms offer different strategies for achieving efficient dataset condensation.

\begin{figure*}[ht]
\label{fig:GEO}
\centering
\subfigure[The training strategies of three different directional paradigms.]{\label{fig:paradigm}
\includegraphics[height=0.26\textheight]{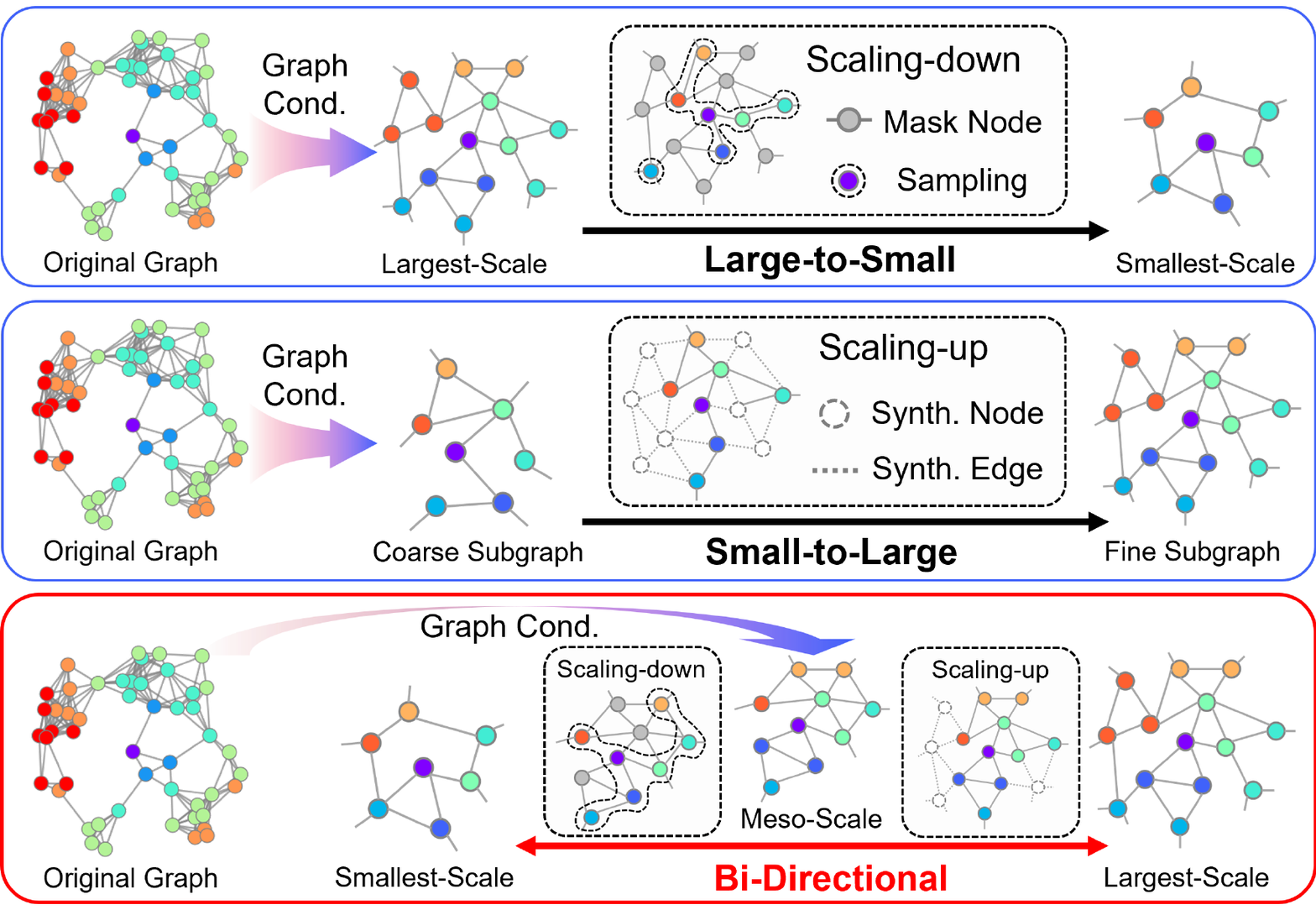}
}
\subfigure[Performance.]{\label{fig:stat}
\includegraphics[height=0.26\textheight]{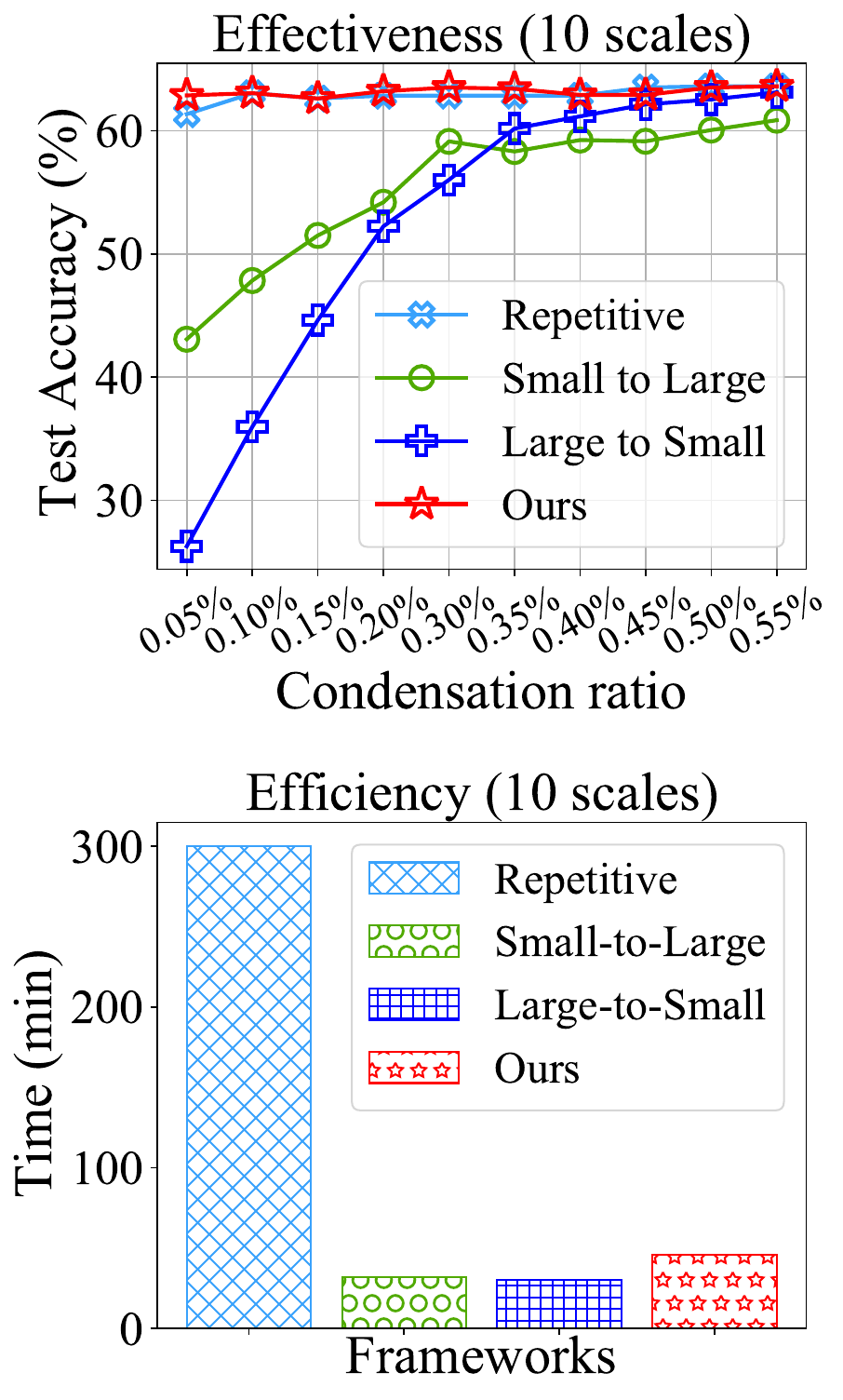}
}
\subfigure[Mutual Information.]{\label{fig:stat2}
\includegraphics[height=0.26\textheight]{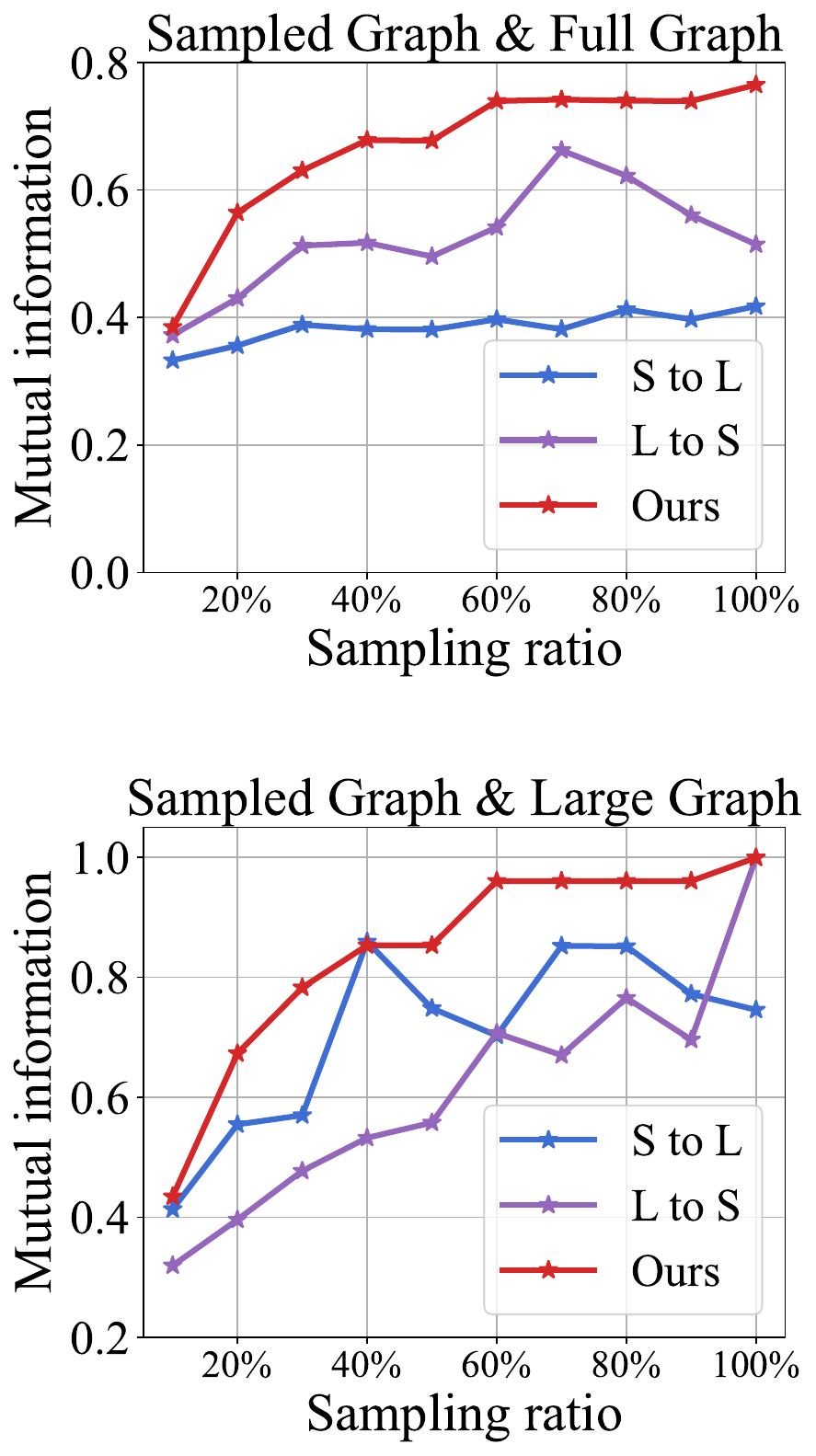}
}
\centering
\vspace{-1em}
\caption{(a) Three different paradigm-specific training strategies: Large-to-Small is trained by mask sampling from a large graph to obtain a small graph. Small-to-Large is trained by considering the small graph as a subgraph expansion of the large graph. Our method obtains meso-scale subgraphs and then trains them separately to both sides;
(b) Condensation performance on each scale (top) and time efficiency (bottom) obtained by training four different multi-scale compression strategies on Ogbn-arxiv using GCond as a backbone method;
(c) The magnitude of mutual information between the different target condensation scales of the three paradigms and the original graph (top), as well as the magnitude of mutual information with the largest scale condensed graph (bottom), where S-to-L represents Small-to-Large and L-to-S represents Large-to-Small.
}
% \vspace{-1em}
\label{fig:GEO}
\end{figure*}

As illustrated in Fig.\ref{fig:stat}, while the Large-to-Small and Small-to-Large paradigms achieve a reduction in time by nearly 10 times. However, there is still a significant effectiveness gap remains compared to the naive re-condensation paradigm at certain scales. Generally, the Large-to-Small approach struggles at relatively small scales, whereas the Small-to-Large approach falters at larger scales. This raises a question of \textit{whether we can combine the advantages of both paradigms to achieve a better balance between efficiency and effectiveness.} In order to better understand the effect of scale variation on the condensation performance, we further analyze the experimental phenomenon and attribute it to two key issues: 
(1)  \textbf{Scaling Down Degradation}: Similar to the subset degradation problem observed in image data~\cite{he2024multisizedatasetcondensation}, graph data suffer from a severe "subgraph degradation problem." When we sample a subgraph from a condensed graph, its performance is much lower than a graph directly condensed to the target scale. In the large-to-small method, the distribution of effective information across a large number of nodes during the initial training process causes a more rapid degradation in performance when the sampling target scale is significantly reduced. (2) \textbf{Scaling Up Collapse}: as demonstrated in Fig.~\ref{fig:stat}, because graph condensation uses fixed neural network model parameters across all scales, the small-to-large method tends to overfit long-tail information to adjust the fine-grained decision boundary at larger scales. This will lead to a large of noisy information during the scale-growing training process, thus limiting the condensation performance at larger scales.

% Inspired by the above, we propose a novel \textit{coarse-to-fine} paradigm to alleviate the above problems.
% The core idea is that, instead of the \textit{large to small} condensation process, our framework is to condense the original graph into the smallest scale and coarse-grained "minimum necessary subgraph", and then "grow" the more detailed larger scale subgraph step by step.
% We aim to prioritize coarse-grained information about the significant features required for the smallest subgraph as much as possible, and then "fill in the details" for the larger subgraphs. 
% Specifically, first, we obtain the "minimum necessary subgraph" based on the significant subgraph mining method (in this paper, we use GNNExplainer), and then supplement the detail sample of the larger subgraph according to different scale requirements.
% Secondly, we dynamically adjust the training trajectories of different scales based on the trajectory matching method of course learning, and re-adjust the loss function based on the guidance of the idea of information bottleneck to solve the problem of parameter redundancy. 
To address these challenges, we first analyze the multi-scale graph condensation through the view of mutual information(Fig.\ref{fig:stat2}). We observe a crucial finding: a mesoscopic scale (meso-scale) exists that balances scaling variation and condensation performance.
Further, we propose starting at a 'meso-scale' and conducting bi-directional scaling to achieve optimal results on both ends.  Inspired by recent research on information bottlenecks in graphs, we first estimate the exact size of the meso-scale and first condense a meso-scale subgraph of size, which is used to retain the maximum amount of valid information while minimizing the scale. Further, we introduce the information bottleneck principle into the optimization of bi-directional condensation to retain more effective information during the scale-changing. As shown in the Fig. \ref{fig:stat}, BiMSGC gains nearly same performance to the naive re-condensation performance at the similar time cost to the Large-to-small and Small-to-Large paradigms. 
We summarize our contributions as follows:
\begin{itemize}
    \item For multi-scale graph dataset condensation, we first explore the limitations of existing work and propose a novel bi-directional multi-scale graph condensation paradigm.
    \item For the problem of retaining valid information in scale variation, we introduce the idea of subgraph condensation information bottleneck to minimize the scaling down degradation and scaling up collapse problems.
    \item Extensive experiments have demonstrated that our BiMSGC outperforms the baselines by a large margin in five datasets. For example, BiMSGC has 20.85 speedup compared with the baselines on Citeseer dataset.
\end{itemize}
\section{Related Works}
\subsection{Graph Dataset Condensation} 
 Graph dataset condensation~\cite{sun2024gcbenchopenunifiedbenchmark} distills the data of graph structures to achieve the same effect of the original graph for graph neural network training.
 GCond~\cite{jin2022graph} distills the graph by matching the model's gradient at each training step, while SFGC~\cite{zheng2023structurefree} extends this to match the entire training trajectory. GDEM~\cite{liu2024graph} offers a different perspective by eliminating spectral domain differences between the synthetic and original graphs through feature basis matching. On the other hand, GCDM~\cite{liu2022graphcondensationreceptivefield}focuses on synthesizing the final compressed graphs by minimizing the distributional differences in the GNN embeddings.

\subsection{Information Bottleneck}
The information bottleneck (IB)~\cite{tishby2000informationbottleneckmethod} originates from rate distortion theory and aims to compress the source variable $X$ into a compact representation $Z$ while retaining the necessary information to predict the target variable $Y$. IB is used to evaluate the rate-distortion trade-off for the source variable and is widely applied in interpretable deep learning theories to enhance robustness~\cite{michael2018on}. In parallel, graph information bottleneck~\cite{wu2020graphinformationbottleneck} has been introduced to characterize and manage the flow of information in graph-structured data.

\section{Problem Formulation and Analysis}
\subsection{Notations}
Given a large graph dataset $\mathbf{G} =(\mathbf{X},\mathbf{A},\mathbf{Y})$, where $\mathbf{X}\in\mathbb{R}^{N\times D}$ denotes the number of N nodes with D-dimensional characteristics, $\mathbf{A}\in\{0,1\}^{N\times N}$ denotes the adjacency matrix of the graph, and $\mathbf{Y}\in\{0,1,...,C-1\}^{N\times 1}$ denotes the label of each node in the C classes. 
The purpose of graph dataset condensation is to condense a large graph $\mathbf{G}$ into a synthetic graph $\mathbf{G}^{\prime}=(\mathbf{X}^{\prime},\mathbf{A}^{\prime},\mathbf{Y}^{\prime})$ while maintaining similar model training results, where $\mathbf{X}^{\prime}\in\mathbb{R}^{N'\times D}$, $\mathbf{A}^{\prime}\in\{0,1\}^{N'\times N'}$, $\mathbf{Y}^{\prime}\in\{0,1,...,C-1\}^{N'\times 1}(N' \ll N)$.
% For multi-scale graph dataset condensation, we need to obtain the condensed datasets at different scales. 
For the condensed graphs at different scales, we use $\mathbf{G}^{\prime}_s$, $\mathbf{G}^{\prime}_m$, $\mathbf{G}^{\prime}_l$ to represent the small-scale, middle-scale, and large-scale condensed dataset respectively.

\subsection{Problem Formulation}
From the perspective of mutual information, the objective of graph dataset condensation can be uniformly expressed as:
\begin{equation}\max I\left(G^{\prime};H(G,Y)\right),\end{equation}
where $I$ denotes the mutual information and $H(G,Y)$ represents the relevant information extracted from the original graph during model training for node classification task. Specifically, the relevant information refers to the gradients or the trajectory of the GNN training process for gradient matching methods, and feature distribution for distribution matching methods.

The task of multi-scale graph dataset condensation involves generating multiple condensed graphs at various scales, each capable of achieving the same training effectiveness as the original dataset. 
Concerning requirements of efficiency, recondensing the graph at every desired scale is impractical, as it would result in  significant time and space inefficiencies. 
A more effective approach is to generate condensed graphs at different scales by selecting subsets from the largest condensed graph.
In other words, $\mathbf{G}^{\prime}_s$, $\mathbf{G}^{\prime}_m$, and $\mathbf{G}^{\prime}_l$ are all subgraphs of $\mathbb{G}^{\prime}$. Our objective then becomes:
\begin{equation}\max I(G_{sub}^{\prime};H(G,Y));\forall G_{sub}^{\prime}\in Sub(G^{\prime}).\end{equation}
where $G_{sub}^{\prime}$ represents a subgraph of the condensed graph $G^{\prime}$, $Sub(G^{\prime})$ denotes the set of all subgraphs of $G^{\prime}$.

% \textbf{Gradient Matching and Eigenbasis Matching}
% GCond et al. ensured the experimental results from the perspective of the training trajectory by matching the gradients of the real and synthetic datasets so that the synthetic dataset reached the neighborhood of the original dataset in terms of the point of model parameters after training. 
% Its objective function is:
% \begin{equation}
% \min_{\mathcal{S}}\mathrm{E}_{\boldsymbol{\theta}_{0}\sim P_{\boldsymbol{\theta}_{0}}}\left[\sum_{t=0}^{T-1}D\left(\nabla_{\boldsymbol{\theta}}\mathcal{L}_{\mathrm{GNN}_{\boldsymbol{\theta}_{t}}}(\mathbf{G}^\mathbb{S}),\nabla_{\boldsymbol{\theta}}\mathcal{L}_{\mathrm{GNN}_{\boldsymbol{\theta}_{t}}}(\mathbf{G}^\mathbb{T})\right)\right]
% \end{equation}

\subsection{Problem Analysis}

% \subsection{Graph Dataset Condensation in the Perspective of Mutual Information}

% That is, we want each subset to retain as much valid gradient information as possible. This process can be naturally translated into an information bottleneck objective. Specifically, the gradient information bottleneck is represented as:
% \begin{equation}\arg\max_{G_{sub}^S}I\left(S_{sub};\nabla_{\theta}^{\prime}\right)-\beta I\left(S_{sub};S\right),\mathrm{s.t.}S_{sub}\in\mathbb{G}_{sub}(S),\end{equation}
% \subsection{The reason for degeneracy and collapse problem }
\textbf{Understanding scaling degeneracy and collapse problems.}
% The biggest challenge, however, is that we want good information retention at every scale possible. This provides a possible explanation for the potential problems with the two multi-scale distillation methods:
Based on experimental observations, we identify two key issues in multi-scale graph dataset condensation: scaling down degradation and scaling up collapse. We further analyze the primary factors contributing to these problems using mutual information theory. Following the approach of GMI \cite{peng2020graphrepresentationlearninggraphical}., we estimate the mutual information $I(G_{sub}^{\prime};G)$ between the sampled subgraph $G_{sub}^{\prime}$ and the original graph $G$ and its mutual information $I(G_{sub}^{\prime};G^{\prime})$ with the condensed graph $G^{\prime}$ at different scales. The results are shown in Fig \ref{fig:stat2}.
%GMI: Graph Representation Learning via Graphical Mutual Information Maximization

\textbf{Large-to-Small}: This method is designed to preserve the mutual information between the large-scale subset $G_l$ and the optimization objective, while progressively enhancing the mutual information for the small-scale subset $G_s$. Since it has a very high value of $I(G_{sub}^{\prime};G)$ with the original graph, this allows it to approach the results of re-condensation at larger scales. However, it is observed that the method yields very low mutual information $I(G_{sub}^{\prime};G^{\prime})$ values at smaller scales, which explains the pronounced subgraph degradation problem in Fig\ref{fig:stat} at target condensation ratio below 0.30\%.

\textbf{Small-to-Large}: This approach begins by preserving mutual information within small-scale graphs $G_s$, gradually introducing new training nodes to expand the scale and incorporate effective mutual information into the larger graph $G_l$. $I(G_{sub}^{\prime};G^{\prime})$ using the small-to-large method is higher than that of the large-to-small method.
It demonstrates that this method effectively maintains a lower bound of mutual information retention at the smallest scale. However, it does not fully address the issue of scale discrepancy. 
It has the smallest mutual information value $I(G_{sub}^{\prime};G)$ among the three condensation methods. This may be due to the introduction of a large number of new nodes in the later stages of training, which makes the node features contain a large amount of noisy information, thus affecting the upper limit of the expressive power of the condensed graph. This explains why the small-to-large method does not test as well as the other methods after the target condensation ratio is greater than 0.35\% in Fig\ref{fig:stat}.

Fortunately, we have identified a mesoscopic-scale condensation graph $G_m$ that effectively mitigates the loss of mutual information caused by scale differences. In addressing the scaling down degradation problem, this approach retains as much mutual information as possible while minimizing scale reduction. Conversely, in the case of scaling up collapse, it reduces the redundancy of extra information that often accompanies scale increases, thereby maintaining a more balanced and efficient condensation process.

% \subsection{Information Bottleneck}
% \bn{Should introduce the basic notations here.}
% Liu~\cite{} found that the case of gradient matching causes the problem of bias in the spectral domain, in which the synthesized map is biased towards high frequencies and the original map is biased towards low frequencies. To solve this problem, they removed the influence of the eigenvalues of the graph structure and transformed it into eigenbase matching:

% \begin{equation}\mathcal{L}_{e}=\sum_{k=1}^{K}\left\|\mathbf{X}^{\top}\mathbf{u}_{k}\mathbf{u}_{k}^{\top}\mathbf{X}-\mathbf{X}^{\prime}{}^{\top}\mathbf{u}_{k}^{\prime}\mathbf{u}_{k}^{\prime}{}^{\top}\mathbf{X}^{\prime}\right\|_{F}^{2},\end{equation}

% \begin{equation}\mathcal{L}_{o}=\left\|\mathbf{U'}_{K}{}^{\top}\mathbf{U'}_{K}-\mathbf{I}_{K}\right\|_{F}^{2}.\end{equation}

% \begin{equation}\mathcal{L}_{d}=\sum_{i=1}^{C}\left(1-\frac{\mathbf{H}_{i}^{\top}\cdot\mathbf{H}_{i}^{\prime}}{||\mathbf{H}_{i}|| ||\mathbf{H}_{i}^{\prime}||}\right)+\sum_{i,j=1}^{C}\frac{\mathbf{H}_{i}^{\top}\cdot\mathbf{H}_{j}^{\prime}}{||\mathbf{H}_{i}|| ||\mathbf{H}_{j}^{\prime}||}.\end{equation}

% \begin{equation}
% \label{Eq:gdem}
% \mathcal{L}_{total}=\alpha\mathcal{L}_{e}+\beta\mathcal{L}_{d}+\gamma\mathcal{L}_{o},\end{equation}

%\input{4_problem}
\section{Methodology}
\begin{figure*}\label{fig:architecture}
    \centering
    \includegraphics[width=0.95\linewidth]{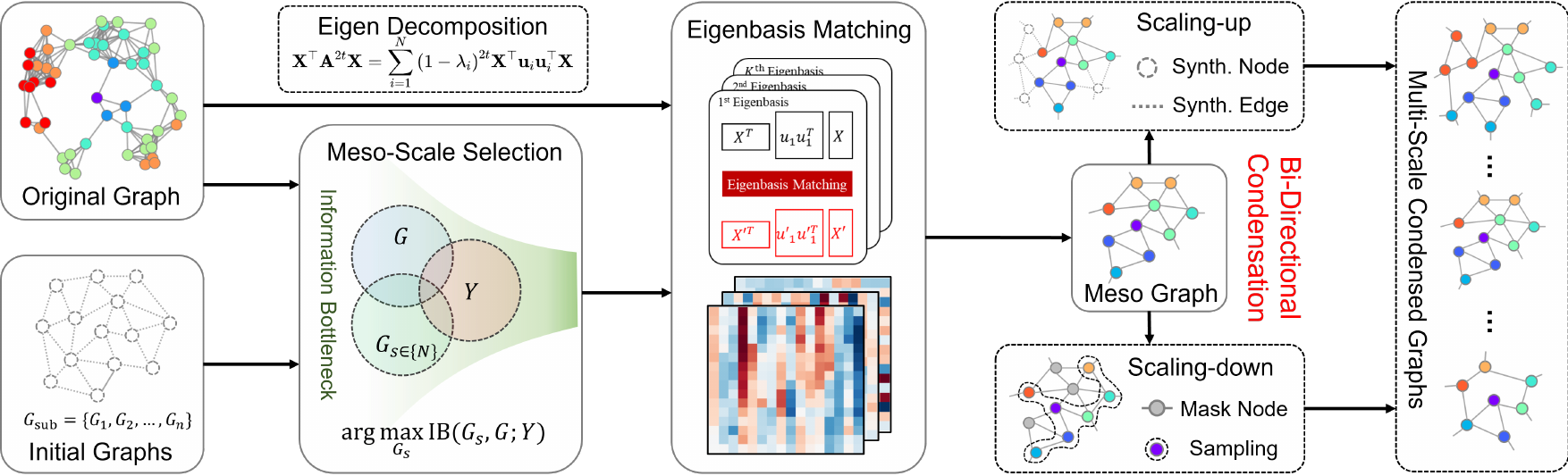}
    \caption{An illustration of BiMSGC  architecture. }
    \label{fig:framework}
\end{figure*}

In this section, we elaborate the proposed \modelname, a novel bi-directional and information bottleneck principle guided multi-scale graph condensation framework. 
Our work mainly consists of three parts. 
First of all, we present how we use the information bottleneck principle to guide meso-scale selection. 
Then, we introduce the bi-directional multi-scale graph dataset condensation optimization method based on information bottleneck. 
Finally, we combine the bi-directional condensation framework with a concrete graph condensation method based on eigenbasis matching.
The architecture is shown in Figure~\ref{fig:framework}, and the overall process of \modelname~is described in Algorithm~\ref{Alg:training}. 

\subsection{Bi-directional Multi-Scale Graph Dataset Condensation Framework}

% An indispensable step of multi-scale graph dataset condensation task is to generate of a subgraph that maximizes the preservation of effective mutual information despite scale differences. 
One indispensable step of multi-scale condensation is to generate a subgraph that best preserves mutual information despite scale difference. 
We start from distilling a meso-scale dataset, and then adjust it to target scales as needed, minimizing the impact of excessive scale differences while maintaining high computational efficiency.

\noindent \textbf{Step 1: Information Bottleneck Guided Meso-Scale Selection.}
 
To achieve a more stable condensation performance across multiple scales, we first distill a 'middle scale' subgraph $G_m$, referred to as \textit{Meso-Scale} subgraph. 
This subgraph captures as much valid information as possible, but minimize redundancies and noise. 
Naturally, we associate it with the subgraph information bottleneck, which aims to compress the data from the original version while preserving key valid information. Specifically, the formulation is as follows:
% \begin{equation}
% \label{middle_goal}
%             S_{m}=\arg\max_{S_{sub}\in S}(I(S_{sub},S_{s})+I(S_{sub},S_{l}))
% \end{equation}
\begin{equation}
 G^{\prime}_{m}=\mathop{\mathrm{argmax}}_{ G^{\prime}_{sub}\in {Sub(G^{\prime})}}I\left(G^{\prime}_{sub};Y^{\prime}\right)-\beta I\left(G^{\prime};G^{\prime}_{sub}\right),
\label{Eq:middle_scale}
\end{equation}
where $\beta$ is a hyperparameter balancing informativeness and compression.

% This process can be initially estimated by the computation of the subgraph IB(cite).For the first part of  Eq \eqref{Eq:middle_scale}, it can obtain a tractable lower bound by substituting the true posterior $p(y^{\prime}|\mathcal{G}_{sub})$ with a variation  approximation $q(y^{\prime}|\mathcal{G}_{sub})$:

% \begin{equation}\begin{aligned}I(Y,\mathcal{G}_{sub})&\geq\int p(y,\mathcal{G}_{sub})\log q_{\phi_{1}}(y|\mathcal{G}_{sub})dy d\mathcal{G}_{sub}\\&\approx\frac{1}{N}\sum_{i=1}^{N}q_{\phi_{1}}(y_{i}|\mathcal{G}_{sub_{i}})\\&=-\mathcal{L}_{cls}(q(y|\mathcal{G}_{sub}),y_{gt}),\end{aligned}\end{equation}

% For the second part $I(S,S_m)$, it is usually estimated by introducing the variational approximation $R_{sub}$ for the marginal distribution $P_{sub}$
% \begin{equation}
% I\left(S_{sub};S\right)\leq\mathbb{E}_{\mathcal{S}}\left[\mathrm{KL}\left(P\left(S_{sub}\right|S\right)\left\|R(S_{sub})\right)\right]
% \end{equation}

% See the appendix for details of the calculation process.
It is worthy to note that this step is only a preliminary estimation rather than a final optimization goal. 
Therefore, it can be approximated using MINE~\cite{belghazi2021minemutualinformationneural} or GIB~\cite{wu2020graphinformationbottleneck}.
% After determining the meso-scale, we first train on that scale and obtain the condensed graph $G_m^{\prime}$. 
In practice, the meso-scale is selected by pre-setting several subgraph scales for calculation and then comparing them.
Having the meso-scale determined, we obtain the condensed graph $G_m^{\prime}$ by training on this scale.

\noindent \textbf{Step 2: Bi-directional Graph Condensation with IB.}
% After obtaining intermediate subgraphs through initial condensation, we further train them in both directions. Further, facing the existing scaling difference problem, we utilize the information bottleneck approach to maximize the retention of key mutual information with scale change. Specifically, according to the information bottleneck principle, information related to the target gradient is retained as the scale changes, while irrelevant information is minimized. This process can be expressed as:

After obtaining the meso-scale subgraph $G_m^{\prime}$ by initial condensation, we further train the synthesized graphs in both directions. Specifically, we view the optimization problem for both training processes uniformly as a variant of the subgraph information bottleneck and call it the \textbf{S}ubgraph \textbf{C}ondensation \textbf{I}nformation \textbf{B}ottleneck(SCIB). It takes the following specific form:
\begin{equation}
\begin{aligned}
\mathop{\mathrm{max}}_{ G^{\prime}_{sub}\in {Sub(G^{\prime})}}I\left(G^{\prime}_{sub};H(G,Y)\right)-\beta I\left(G^{\prime};G^{\prime}_{sub}\right),
\end{aligned}
\label{ib2}
\end{equation}

Considering the difficulty to compute and optimize the mutual information directly, VIB~\cite{alemi2019deepvariationalinformationbottleneck} optimized the objective by using a variational approach to compute the under-error session of the information bottleneck.

To be specific, for the first part $I\left(G^{\prime}_{sub};H(G,Y)\right)$, 
we replace $ P(H(G,Y)|G_{sub}^{\prime})$ with a parameterized variational approximation $ Q(H(G,Y)|G_{sub}^{\prime})$ and infer its approximate lower bound:
\begin{equation}
I(G_{sub}^{\prime};H(G,Y))\geq\mathbb{E}\left[\log Q(H(G,Y)|G_{sub}^{\prime})\right],
\end{equation}
Simultaneously, for the second part $I(G^{\prime},G_{sub}^{\prime})$, we estimate it by introducing the variational approximation $R$ for the marginal distribution $P$, following common practice. With Kullback-Leibler (KL) divergence, the mutual information can be approximated as:
\begin{equation}
\begin{aligned}
I\left(G_{sub}^{\prime};G^{\prime}\right)&\leq\mathbb{E}\left[\mathrm{KL}\left(P\left(G_{sub}^{\prime}\right|G^{\prime}\right)\left\|R(G_{sub}^{\prime})\right)\right]\\
& = \mathbb{E}\left[f(G_{sub}^{\prime},G^{\prime})\right],
\end{aligned}
\end{equation}

By combining these two equations together, we derive the optimization objective $L_{SCIB}$:
\begin{equation}
\begin{aligned}
L_{SCIB}(G_{sub}^{\prime};G^{\prime})&=\mathbb{E}\left[\log Q(H(G,Y)|G_{sub}^{\prime})\right]\\&-\beta\mathbb{E}\left[f(G_{sub}^{\prime},G^{\prime})\right].
\end{aligned}
\label{Eq:ibsmall}
\end{equation}

For condensation from meso-scale to smaller scales, we consider the small-scale graphs $G_l^{\prime}$ to be subgraphs of the meso-scale graph $G_m^{\prime}$. In this case, its optimization objective would be $L_{SCIB}(G_{l}^{\prime};G_m^{\prime})$.
Similarly, for the process of expanding from meso-scale to large scales, we treat $G_m^{\prime}$ as a subgraph of large-scale graphs, with  $L_{SCIB}(G_{m}^{\prime};G^{\prime})$ as the corresponding optimization. The detailed derivation is reported in the Appendix.

\noindent \textbf{Step 3: Instantiation of Multi-Scale Condensed Graph.}

We instantiate the distribution $ Q$  assigning importance score to each node's impact on the training loss function, and then define the distribution $R$ as a Bernoulli distribution with parameter $\theta$. For $P\left(G_{sub}|G\right)$, we assign relevant scores to different scales by continuously adjusting global masks. While transitioning from meso to smaller scales, we gradually increase the global masks to reduce the training scale. Conversely, the global masks are decreased while expanding the training scale.
\subsection{Application on Graph Condensation}
In this section, we integrate the above framework with specific techniques for graph condensation. 
Theoretically, our approach can be combined with any graph condensation method to represent $H(G,Y)$. 
In order to retain more important information, we use eigenbasis matching method, which leverages key structural information along with gradient information. 

First, we compute the Laplacian matrix $\mathbf{L}$ from the adjacency matrix $\mathbf{A}$ of the original graph and decompose it as $\mathbf{L}=\mathbf{U}\mathbf{\Lambda}\mathbf{U}^{\top}=\sum_{i=1}^{N}\lambda_{i}\mathbf{u}_{i}\mathbf{u}_{i}^{\top}$. 
We then initialize the corresponding eigenbasis $\mathbf{U}_{K}^{\prime} = [\mathbf{u}_{1}^{\prime},\cdots,\mathbf{u}_{N^{\prime}}^{\prime}] \in \mathbb{R}^{N^{\prime}\times K}$, where $K$ is a hyperparameter.

For a simple GNN model, the loss function can be expressed as:

\begin{equation}
\begin{aligned}
\mathcal{L}_{GM}&=\|\nabla_{\mathbf{W}}-\nabla_{\mathbf{W}}^{\prime}\|_{F}^{2}\\
&\leq\|\mathbf{W}\|\|\mathbf{X}^{\top}\mathbf{A}^{2t}\mathbf{X}-\mathbf{X}^{\prime}{}^{\top}\mathbf{A}^{\prime}{}^{2t}\mathbf{X}^{\prime}\|
\\&\quad +\|\mathbf{X}^{\top}\mathbf{A}\mathbf{Y}-\mathbf{X}^{\prime}{}^{\top}\mathbf{A}^{\prime}\mathbf{Y}^{\prime}\|
\\&\approx \|\mathbf{W}\|  \underbrace{  \textstyle\sum_{i=1}^{N  ^{\prime}}(\mathbf{X}^\top\mathbf{u}_i\mathbf{u}_i^\top\mathbf{X}-\mathbf{X}^{\prime}{}^{\top}\mathbf{u}_{i }^{\prime}\mathbf{u}_{i}^{\prime}{}^{\top}\mathbf{X}^{\prime})}_{\mathcal{L}_e}
\\&\quad+\underbrace{\|\mathbf{X}^{\top}\mathbf{A}\mathbf{Y}-\mathbf{X}^{\prime}{}^{\top}\mathbf{A}^{\prime}\mathbf{Y}^{\prime}\|}_{\mathcal{L}_o},
\end{aligned}
\end{equation} 

Due to the orthogonality of eigenvector matrices, the representation space is constrained by a regularization loss:
\begin{equation}\mathcal{L}_{o}=\left\|\mathbf{U'}_{K}{}^{\top}\mathbf{U'}_{K}-\mathbf{I}_{K}\right\|_{F}^{2},\end{equation}

Generally, the optimization objective for eigenbasis matching is:
\begin{equation}
\label{Eq:train_loss}
\mathcal{L}_{total}=\alpha\mathcal{L}_{e}+\beta\mathcal{L}_{d}+\gamma\mathcal{L}_{o}.\end{equation}
where $\alpha$, $\beta$, $\gamma$ are hyperparemeters.

As for condensing multi-scale graphs, we first take the eigenvectors corresponding to the first $M$ eigenvalues and distill a meso-scale subgraph utilizing the equation. 
After that, we still match with the first $M$ feature values for further training of small-scale graphs from meso-scale ones. For going from meso to large scales, the matching is shifted to the feature vectors corresponding to the first $N^{\prime}$ feature values.

\begin{algorithm}[t]
\caption{Bi-directional Condensation Algorithm}
\begin{algorithmic}
        \STATE {\bfseries Input:} Original Graph $\mathbf{G}=\{\mathbf{X},\mathbf{A}\}$; Number of training meso-scale epochs $E_1$; Number of training bi-directional scale epochs $E_2$, the learning rate $\eta$.\\
        \STATE {\bfseries Parameter:} The relevant parameters $\theta$ about synthesized graph $G^{\prime}$.\\
        \STATE {\bfseries Output:}  The condensed graph $G^{\prime}$.\\
        \STATE Init a meso-scale $G_M$ using Eq~\eqref{Eq:middle_scale}.\\
        \FOR{$e=1$ {\bfseries to} $E_1$}
            \STATE  Train $G_M$ using Eq~\eqref{Eq:train_loss}.\\
             \STATE Update $\theta_{m} \gets \theta_m-\eta \nabla \theta_m$.
        \ENDFOR
        \FOR{$e=1$ {\bfseries to} $E_2$}
            \STATE Train $G_s$ using Eq~\eqref{Eq:train_loss} and Eq~\eqref{Eq:ibsmall};\\
            \STATE Train $G_l$ using Eq~\eqref{Eq:train_loss} and Eq~\eqref{Eq:ibsmall};\\
           \STATE  Update $\theta \gets \theta-\eta \nabla \theta$.
        \ENDFOR
    \end{algorithmic} 
\label{Alg:training}
\end{algorithm}

\subsection{Complexity Analysis} 
First, we decompose the $K$ largest eigenvalues of the original image, of which the time complexity is $\mathcal{O}(KN^2)$. 
Complexity of calculating the loss function \eqref{Eq:train_loss} is $\mathcal{O}(KN^{\prime}d+Kd^{2}+KN^{\prime2})$. 
Since the training process consists of two phases, the time complexity of the whole process is a total of $\mathcal{O}(2(KN^{\prime}d+Kd^{2}+KN^{\prime2}))$. 
However, since the first stage training already contains some of the optimization objectives for the second stage training, the corresponding training time can be reduced. 
The actual time complexity of our method can be approximated as $\mathcal{O}(KN^{\prime}d+Kd^{2}+KN^{\prime2})$, while re-condensation requires to train at each scale separately. 
To sum up, it has a complexity of $\mathcal{O}(N^{\prime}(KN^{\prime}d+Kd^{2}+KN^{\prime2}))$ which is an order of magnitude higher in complexity than our method.
\section{Experiment}
\subsection{Experimental Setup}

\textbf{Datasets.}
To evaluate the performance of our \modelname~\footnote{\texttt{{https://github.com/RingBDStack/BiMSGC}}.}, we choose five node classification benchmark graphs, including three transductive graphs, Cora, Citeseer~\cite{kipf2017semisupervised}, Ogbn-Arxiv~\cite{hu2021ogblsclargescalechallengemachine} and two inductive graphs, Flickr and Reddit~\cite{zeng2020graphsaintgraphsamplingbased}. 
% These datasets vary in sizes with number of nodes ranging from 2,708 to 232,965, thus providing a comprehensive evaluation.

% \begin{table}[t]
% \centering
% \resizebox{\linewidth}{!}{
% \begin{tabular}{lrrrr}
% \toprule
% \textbf{Dataset} & \textbf{\#Nodes} & \textbf{\#Edges} & \textbf{\#Classes} & \textbf{\#Features} \\
% \midrule
% Citeseer     & 3,327   & 4,732    & 6    & 3,703  \\
% Cora         & 2,708   & 5,429    & 7    & 1,433  \\
% Ogbn-Arxiv   & 169,343 & 1,166,243 & 40   & 128    \\
% % \midrule
% Flickr       & 89,250  & 899,756  & 7    & 500    \\
% Reddit       & 232,965 & 57,307,946 & 210  & 602    \\
% \bottomrule
% \end{tabular}}
% \caption{Statistics of Datasets}\label{table:datasets}
% \end{table}

\noindent\textbf{Baselines.}
Based on GC-Bench~\cite{sun2024gcbenchopenunifiedbenchmark}~\footnote{\texttt{{https://github.com/RingBDStack/GC-Bench}}.}, we select gradients-based methods: GCond~\cite{jin2022graph}, SGDD~\cite{yang2023does}, EXGC~\cite{fang2024exgc}; trajectory-based methods:
SFGC~\cite{zheng2023structurefree},GEOM~\cite{zhang2024navigating};
other method: GCDM~\cite{liu2022graphcondensationreceptivefield}, GC-SNTK~\cite{liu2022graphcondensationreceptivefield}, GDEM~\cite{liu2024graph}.

\begin{table*}[t]
% gc-sntk omitted for out of time on reddit dataset
\centering
\setlength{\tabcolsep}{1.5mm}{
\footnotesize
\begin{tabular}{c|c|cccccccc|c|c}
\toprule
\multirow{2}{*}{\textbf{Dataset}} & \multirow{2}{*}{\begin{tabular}[c]{@{}c@{}}Reduction \\ rate(\%)\end{tabular}} & \multicolumn{9}{c|}{\textbf{Condensation Methods}} & \multirow{2}{*}{\begin{tabular}[c]{@{}c@{}}Whole \\ Dataset\end{tabular}} \\ \cline{3-11} 
 & & \raisebox{-1mm}{\textbf{GCond}} & \raisebox{-1mm}{\textbf{SFGC}} & \raisebox{-1mm}{\textbf{SGDD}} & \raisebox{-1mm}{\textbf{EXGC}} & \raisebox{-1mm}{\textbf{GC-SNTK}} & \raisebox{-1mm}{\textbf{GCDM}} & \raisebox{-1mm}{\textbf{GDEM}} & \raisebox{-1mm}{\textbf{GEOM}} & \raisebox{-1mm}{\textbf{Ours}} & \\ 
 \midrule
\multirow{4}[2]{*}{\textbf{Cora}} 
& 0.50  & 58.0$_{\pm 5.4}$ & 66.9$_{\pm 7.0}$ & 61.6$_{\pm 6.3}$ & 61.5$_{\pm 7.2}$ & 47.4$_{\pm 13.3}$ & 53.2$_{\pm 2.7}$ & \underline{68.7}$_{\pm 3.6}$ & 39.9$_{\pm 10.4}$ & \textbf{78.2}$_{\pm 1.3}$ 
% avg
& \multirow{4}{*}{80.9$_{\pm 0.1}$}\\
& 1.00 & 73.4$_{\pm 4.6}$ & 70.3$_{\pm 0.6}$ & \underline{74.4}$_{\pm 2.0}$ & 71.9$_{\pm 2.5}$ & 65.4$_{\pm 10.0}$ & 61.2$_{\pm 4.8}$ & 69.9$_{\pm 1.7}$& 66.0$_{\pm 6.6}$ & \textbf{80.5}$_{\pm 1.0}$ &\\
& 1.50 & 79.2$_{\pm 1.2}$ & \underline{79.3}$_{\pm 0.5}$ & 75.5$_{\pm 2.3}$ & 77.9$_{\pm 1.2}$ & 71.6$_{\pm 4.9}$ & 56.0$_{\pm 3.0}$ & 71.1$_{\pm 2.9}$ & 77.6$_{\pm 2.1}$ & \textbf{79.5}$_{\pm 1.4}$ & \\
& 2.00 & 80.0$_{\pm 0.3}$ & 80.8$_{\pm 0.4}$ & 77.9$_{\pm 1.0}$ & 79.8$_{\pm 0.4}$ & 74.8$_{\pm 2.6}$ & 79.0$_{\pm 3.0}$ & 74.1$_{\pm 2.9}$ & \textbf{83.6}$_{\pm 0.2}$ & \underline{80.9}$_{\pm 0.6}$ & \\
\midrule  
\multirow{4}[2]{*}{\textbf{CiteSeer}} 
& 0.50 & 51.9$_{\pm 1.0}$ & 33.3$_{\pm 2.0}$ & 41.7$_{\pm 9.1}$ & 45.9$_{\pm 8.2}$ & 40.3$_{\pm 13.0}$ & 51.8$_{\pm 2.6}$ & \underline{68.1}$_{\pm 2.5}$& 32.3$_{\pm 5.9}$ & \textbf{73.7}$_{\pm 0.9}$ 
% avg
& \multirow{4}{*}{71.1$_{\pm 0.1}$}\\
& 1.00 & 53.4$_{\pm 1.4}$ & 32.9$_{\pm 1.8}$ & 52.2$_{\pm 7.2}$ & 59.4$_{\pm 7.1}$ & 47.8$_{\pm 8.2}$ & 58.2$_{\pm 0.2}$ & \underline{70.9}$_{\pm 1.0}$ & 47.5$_{\pm 8.0}$ & \textbf{73.6}$_{\pm 0.6}$ &\\
& 1.50 & 62.6$_{\pm 0.7}$ & 54.5$_{\pm 0.6}$ & 51.0$_{\pm 7.0}$ & 61.5$_{\pm 8.1}$ & 55.6$_{\pm 8.9}$ & 62.9$_{\pm 0.8}$ & \underline{71.8}$_{\pm 1.1}$ & 66.0$_{\pm 4.2}$ & \textbf{73.6}$_{\pm 1.8}$ & \\
& 2.00 & 69.8$_{\pm 0.3}$ & 60.2$_{\pm 0.6}$ & 70.9$_{\pm 0.3}$ & 71.0$_{\pm 0.6}$ & 65.4$_{\pm 5.1}$ & 65.7$_{\pm 1.1}$ & 72.6$_{\pm 0.3}$ & \textbf{74.3}$_{\pm 0.1}$ & \underline{73.6}$_{\pm 0.2}$ & \\
\midrule
\multirow{4}[2]{*}{\textbf{Ogbn-Arxiv}} 
& 0.05 & 48.8$_{\pm 4.0}$ & 45.3$_{\pm 3.0}$ & 52.2$_{\pm 0.9}$ & 44.1$_{\pm 3.9}$ & 25.1$_{\pm 5.6}$ &  51.9$_{\pm 1.7}$ & \underline{61.0}$_{\pm 1.9}$ & 44.9$_{\pm 4.4}$ & \textbf{62.7}$_{\pm 0.4}$ 
% avg
& \multirow{4}{*}{71.8$_{\pm 0.1}$}\\
& 0.10 & 54.0$_{\pm 0.7}$ & 52.6$_{\pm 0.8}$ & 60.2$_{\pm 0.3}$ & 51.0$_{\pm 3.0}$ & 25.1$_{\pm 5.6}$ &  60.3$_{\pm 2.9}$ &\underline{60.5}$_{\pm 2.2}$  & 50.5$_{\pm 3.3}$ & \textbf{63.2}$_{\pm 0.2}$ &\\
& 0.30 & 60.2$_{\pm 1.1}$ & 62.3$_{\pm 1.2}$ & \underline{63.4}$_{\pm 0.1}$ & 59.2$_{\pm 1.9}$ & 46.1$_{\pm 3.8}$ &  62.4$_{\pm 2.7}$ & 60.6$_{\pm 1.4}$ & \textbf{63.8}$_{\pm 0.9}$ & 62.7$_{\pm 0.2}$ & \\
& 0.50 & \underline{64.5}$_{\pm 0.1}$ & \textbf{66.2}$_{\pm 0.4}$ & 63.4$_{\pm 0.3}$ & 62.8$_{\pm 0.5}$ & 55.4$_{\pm 1.2}$ &  64.2$_{\pm 2.9}$ & 62.1$_{\pm 1.0}$ & 63.8$_{\pm 0.2}$ & 63.7$_{\pm 0.2}$ & \\
% \multirow{4}[2]{*}{\textbf{Cora}}
\midrule
\multirow{4}[2]{*}{\textbf{Flickr}}
& 0.10 & \underline{46.2}$_{\pm 2.1}$ & 44.1$_{\pm 1.5}$ & 44.0$_{\pm 2.1}$ & 42.2$_{\pm 1.2}$ & 28.3$_{\pm 5.7}$ & 32.9$_{\pm 2.5}$ & 42.6$_{\pm 2.5}$ & 43.0$_{\pm 1.7}$ & \textbf{50.3}$_{\pm 0.5}$
% avg
& \multirow{4}{*}{47.2$_{\pm 0.1}$}\\
& 0.30 & \underline{46.8}$_{\pm 0.1}$ & 43.7$_{\pm 0.7}$ & 45.6$_{\pm 0.2}$ & 44.6$_{\pm 0.9}$ & 29.1$_{\pm 3.9}$ & 36.2$_{\pm 0.8}$ & 45.4$_{\pm 2.6}$  & 45.6$_{\pm 0.5}$ & \textbf{50.4}$_{\pm 0.1}$ \\
& 0.70 & 47.1$_{\pm 0.2}$ & 44.3$_{\pm 0.3}$ & \underline{48.0}$_{\pm 0.2}$ & 46.3$_{\pm 0.4}$ & 32.2$_{\pm 6.0}$ & 36.2$_{\pm 1.3}$ & 46.3$_{\pm 2.4}$  & 46.7$_{\pm 0.2}$ & \textbf{50.6}$_{\pm 0.3}$ \\
& 1.00 & 47.1$_{\pm 0.1}$ & 44.2$_{\pm 0.1}$ & 47.9$_{\pm 0.1}$ & 46.9$_{\pm 0.1}$ & 31.2$_{\pm 3.3}$ & 35.9$_{\pm 0.5}$ & \underline{49.4}$_{\pm 3.4}$ & 47.3$_{\pm 0.1}$ & \textbf{50.6}$_{\pm 0.1}$ \\
\midrule
\multirow{4}[2]{*}{\textbf{Reddit}}
& 0.05 & 73.0$_{\pm 5.2}$ & 64.4$_{\pm 0.8}$ & 81.0$_{\pm 0.9}$ & 71.4$_{\pm 2.9}$ & - &  78.0$_{\pm 3.4}$ & \underline{83.6}$_{\pm 0.9}$ & 69.0$_{\pm 5.7}$ & \textbf{93.3}$_{\pm 0.7}$ 
% avg
& \multirow{4}{*}{93.9$_{\pm 0.1}$}\\
& 0.10 & 84.2$_{\pm 1.5}$ & 67.6$_{\pm 0.2}$ & 82.7$_{\pm 0.2}$ & 82.7$_{\pm 2.1}$ & - & \underline{88.9}$_{\pm 1.6}$  & 86.6$_{\pm 1.0}$ & 83.9$_{\pm 1.2}$ & \textbf{93.6}$_{\pm 0.2}$\\
& 0.15 & 87.4$_{\pm 0.8}$ & 80.3$_{\pm 0.3}$ & 81.9$_{\pm 0.7}$ & 87.1$_{\pm 1.0}$ & - & 85.9$_{\pm 0.4}$ & 86.5$_{\pm 0.3}$  & \underline{88.9}$_{\pm 0.6}$ & \textbf{92.9}$_{\pm 0.1}$\\
& 0.20 & 90.9$_{\pm 0.2}$ & 84.6$_{\pm 1.9}$ & 88.5$_{\pm 0.2}$ & 89.4$_{\pm 0.3}$ & - & 89.2$_{\pm 0.5}$  & \textbf{93.1}$_{\pm 0.1}$ & 91.5$_{\pm 0.1}$ & \underline{93.1}$_{\pm 0.1}$ \\
\bottomrule
\end{tabular}}
\caption{Node classification performance of different condensation methods,. (Result: average score $\pm$ standard deviation. \textbf{Bold}: best; \underline{Underline}: runner-up. -: cuda out of memory.)}
\label{table:performance_gcn}
\end{table*}

\noindent\textbf{Settings and Hyperparameters.}
% To eliminate randomness, we run the distillation steps 5 times and yield 5 synthetic graphs. 
We followed the default values of parameters for baselines.
% , and set the learning rate of our model to 0.0001. 
% We ran 2000 epochs for testing. 
All models were trained and tested on a single Nvidia A800 80GB GPU.
The remaining details are given in the Appendix.

 \begin{table*}[t]
\centering
\setlength{\tabcolsep}{1.5mm}{
\footnotesize
\begin{tabular}{c|c|cccccccc|c}
\toprule
\multirow{2}{*}{\textbf{Dataset}} & \multirow{2}{*}{\textbf{Models}} & \multicolumn{9}{c}{\textbf{Condensation Methods}} \\ \cline{3-11} 
 & & \raisebox{-1mm}{\textbf{GCond}} & \raisebox{-1mm}{\textbf{SFGC}} & \raisebox{-1mm}{\textbf{SGDD}} & \raisebox{-1mm}{\textbf{EXGC}} & \raisebox{-1mm}{\textbf{GC-SNTK}} & \raisebox{-1mm}{\textbf{GCDM}} & \raisebox{-1mm}{\textbf{GDEM}} & \raisebox{-1mm}{\textbf{GEOM}} & \raisebox{-1mm}{\textbf{Ours}} \\
\midrule
\multirow{5}{*}{Citeseer} 
& GCN     & 59.4$_{\pm 8.4}$ & 45.2$_{\pm 14.2}$ & 54.0$_{\pm 12.2}$ & 59.5$_{\pm 10.4}$ & 52.3$_{\pm 10.8}$ & 59.7$_{\pm 6.1}$ & \underline{70.9}$_{\pm 2.0}$ & 55.0$_{\pm 18.8}$ & \textbf{73.6}$_{\pm 0.1}$ \\
& MLP     & 57.9$_{\pm 6.5}$ & 37.8$_{\pm 18.1}$ & 55.8$_{\pm 4.1}$ & 60.3$_{\pm 1.8}$ & 52.3$_{\pm 12.8}$ & 50.6$_{\pm 12.7}$ & \underline{72.8}$_{\pm 5.0}$ & 55.0$_{\pm 18.9}$ & \textbf{73.2}$_{\pm 0.4}$ \\
& SGC     & 50.6$_{\pm 6.9}$ & 40.1$_{\pm 14.1}$ & 55.0$_{\pm 10.4}$ & 59.3$_{\pm 12.1}$ & 53.2$_{\pm 15.0}$ & 28.2$_{\pm 3.0}$ & \underline{72.3}$_{\pm 0.1}$ & 55.5$_{\pm 18.2}$ & \textbf{73.4}$_{\pm 0.1}$ \\
& APPNP   & 52.0$_{\pm 19.1}$ & 35.3$_{\pm 10.1}$ & 56.2$_{\pm 16.4}$ & 55.3$_{\pm 16.3}$ & 38.8$_{\pm 13.9}$ & 62.5$_{\pm 3.8}$ & \underline{71.5}$_{\pm 0.4}$ & 43.6$_{\pm 15.3}$ & \textbf{73.3}$_{\pm 0.3}$ \\
& ChebNet & 47.3$_{\pm 8.5}$ & 49.8$_{\pm 15.0}$ & 55.5$_{\pm 11.1}$ & 64.3$_{\pm 3.8}$ & 60.5$_{\pm 10.6}$ & 37.4$_{\pm 5.5}$ & \underline{71.9}$_{\pm 0.8}$ & 57.7$_{\pm 15.6}$ & \textbf{73.2}$_{\pm 0.2}$ \\
\midrule
\multirow{5}{*}{Cora} 
& GCN     & 72.6$_{\pm 10.2}$ & \underline{74.3}$_{\pm 6.8}$ & 72.3$_{\pm 7.3}$ & 72.8$_{\pm 8.3}$ & 64.8$_{\pm 12.3}$ & 62.4$_{\pm 11.6}$ & 70.9$_{\pm 2.3}$ & 66.8$_{\pm 19.4}$ & \textbf{79.8}$_{\pm 1.2}$ \\
& MLP     & 70.6$_{\pm 2.4}$ & \underline{77.5}$_{\pm 4.6}$ & 71.9$_{\pm 7.5}$ & 71.1$_{\pm 5.6}$ & 63.4$_{\pm 14.3}$ & 52.6$_{\pm 9.0}$ & 68.8$_{\pm 6.9}$ & 66.8$_{\pm 19.4}$ & \textbf{79.1}$_{\pm 0.6}$ \\
& SGC     & 64.9$_{\pm 16.2}$ & \underline{76.6}$_{\pm 3.4}$ & 71.0$_{\pm 9.1}$ & 71.0$_{\pm 10.5}$ & 65.4$_{\pm 7.1}$ & 33.8$_{\pm 15.2}$ & 74.5$_{\pm 0.6}$ & 66.5$_{\pm 19.3}$ & \textbf{79.7}$_{\pm 0.9}$ \\
& APPNP   & 62.6$_{\pm 15.2}$ & \underline{77.1}$_{\pm 3.6}$ & 69.8$_{\pm 11.0}$ & 71.5$_{\pm 9.8}$ & 39.6$_{\pm 11.9}$ & 68.0$_{\pm 11.8}$ & 66.1$_{\pm 10.5}$ & 45.6$_{\pm 14.8}$ & \textbf{79.9}$_{\pm 0.6}$ \\
& ChebNet & 69.1$_{\pm 7.3}$ & \textbf{75.7}$_{\pm 4.2}$ & 73.00$_{\pm 5.2}$ & 72.9$_{\pm 5.9}$ & 67.7$_{\pm 9.2}$ & 39.8$_{\pm 6.9}$ & 68.0$_{\pm 9.0}$ & 67.00$_{\pm 15.2}$ & \underline{75.3}$_{\pm 0.8}$ \\
\midrule
\multirow{5}{*}{Ogbn-Arxiv} 
& GCN      & 56.9$_{\pm 6.9}$  & 56.6$_{\pm 9.4}$  & 59.8$_{\pm 5.3}$  & 54.3$_{\pm 8.3}$  & 38.0$_{\pm 15.2}$ & 59.7$_{\pm 5.4}$  & \underline{61.1}$_{\pm 0.7}$  & 57.0$_{\pm 11.1}$ & \textbf{63.1}$_{\pm 0.4}$ \\
& MLP      & 54.8$_{\pm 5.9}$  & 57.4$_{\pm 8.4}$  & \underline{60.4}$_{\pm 4.1}$  & 54.0$_{\pm 8.3}$  & 37.9$_{\pm 12.5}$ & 50.3$_{\pm 3.7}$  & 58.4$_{\pm 1.6}$  & 57.0$_{\pm 11.1}$ & \textbf{62.4}$_{\pm 0.8}$ \\
& SGC      & 55.8$_{\pm 8.8}$  & 56.1$_{\pm 9.2}$  & \underline{61.7}$_{\pm 2.5}$  & 53.2$_{\pm 9.9}$  & 40.1$_{\pm 16.7}$ & 58.1$_{\pm 5.3}$  & \textbf{63.2}$_{\pm 0.1}$  & 54.2$_{\pm 11.2}$ & 60.2$_{\pm 2.2}$ \\
& APPNP    & 57.3$_{\pm 7.3}$  & 54.8$_{\pm 7.3}$  & 59.6$_{\pm 3.3}$  & 55.2$_{\pm 7.4}$  & 36.0$_{\pm 17.9}$ & 54.4$_{\pm 3.0}$  & \underline{60.5}$_{\pm 1.7}$  & 53.1$_{\pm 10.4}$ & \textbf{61.8}$_{\pm 1.6}$ \\
& ChebNet  & 50.8$_{\pm 8.2}$  & 52.4$_{\pm 8.6}$  & 51.7$_{\pm 5.5}$  & 48.4$_{\pm 8.9}$  & 22.9$_{\pm 11.6}$ & 39.2$_{\pm 3.6}$  & \textbf{57.4}$_{\pm 1.0}$  & 50.4$_{\pm 11.0}$ & \underline{56.9}$_{\pm 0.2}$ \\
\midrule
\multirow{5}{*}{Flickr}     
& GCN      & \underline{46.8}$_{\pm 0.4}$  & 44.1$_{\pm 0.2}$  & 46.4$_{\pm 1.9}$  & 45.0$_{\pm 2.1}$  & 30.2$_{\pm 1.7}$  & 35.3$_{\pm 1.5}$  & 45.9$_{\pm 2.8}$  & 45.7$_{\pm 1.8}$  & \textbf{50.5}$_{\pm 0.1}$ \\
& MLP      & 42.9$_{\pm 0.5}$  & 41.0$_{\pm 0.6}$  & 43.1$_{\pm 1.2}$  & 41.9$_{\pm 1.2}$  & 25.4$_{\pm 3.3}$  & 42.8$_{\pm 1.4}$  & 39.4$_{\pm 0.6}$  & \underline{45.7}$_{\pm 1.8}$  & \textbf{49.7}$_{\pm 0.4}$ \\
& SGC      & \underline{46.6}$_{\pm 0.3}$  & 45.0$_{\pm 0.9}$  & 45.5$_{\pm 2.3}$  & 44.9$_{\pm 1.6}$  & 32.4$_{\pm 3.9}$  & 34.1$_{\pm 0.9}$  & 39.8$_{\pm 0.2}$  & 45.3$_{\pm 1.7}$  & \textbf{49.4}$_{\pm 0.3}$ \\
& APPNP    & 31.1$_{\pm 3.9}$  & 36.0$_{\pm 2.3}$  & 41.8$_{\pm 2.6}$  & 31.6$_{\pm 3.5}$  & 21.6$_{\pm 3.4}$  & \underline{42.8}$_{\pm 1.4}$  & 31.5$_{\pm 0.1}$  & 36.8$_{\pm 3.4}$  & \textbf{49.9}$_{\pm 0.1}$ \\
& ChebNet  & 42.1$_{\pm 1.7}$  & 40.8$_{\pm 0.5}$  & 41.6$_{\pm 1.0}$  & 40.8$_{\pm 1.0}$  & 27.1$_{\pm 1.8}$  & 41.6$_{\pm 1.0}$  & 38.2$_{\pm 0.6}$  & \underline{42.5}$_{\pm 3.0}$  & \textbf{46.0}$_{\pm 0.5}$ \\
\midrule

\multirow{5}{*}{Reddit} 
& GCN      & 83.9$_{\pm 7.7}$  & 74.2$_{\pm 9.7}$  & 83.5$_{\pm 3.3}$  & 82.7$_{\pm 7.9}$  & - & 85.5$_{\pm 5.2}$  & \underline{87.4}$_{\pm 4.0}$  & 83.3$_{\pm 10.0}$ & \textbf{93.2}$_{\pm 0.3}$ \\
& MLP      & 49.5$_{\pm 6.4}$  & 57.4$_{\pm 8.4}$  & 50.3$_{\pm 6.7}$  & 42.8$_{\pm 4.1}$  & - & 77.7$_{\pm 8.7}$  & 54.8$_{\pm 7.4}$  & \underline{83.3}$_{\pm 10.0}$ & \textbf{93.2}$_{\pm 0.1}$ \\
& SGC      & \underline{86.5}$_{\pm 6.0}$  & 56.1$_{\pm 9.2}$  & 84.0$_{\pm 2.9}$  & 83.3$_{\pm 8.2}$  & - & 83.8$_{\pm 3.1}$  & 86.5$_{\pm 2.5}$  & 82.4$_{\pm 10.4}$ & \textbf{89.4}$_{\pm 0.6}$ \\
& APPNP    & 81.9$_{\pm 6.8}$  & 66.2$_{\pm 12.8}$ & \underline{82.8}$_{\pm 7.1}$  & 81.9$_{\pm 7.1}$  & - & 82.6$_{\pm 6.7}$  & 77.7$_{\pm 14.1}$ & 81.7$_{\pm 9.2}$  & \textbf{92.3}$_{\pm 0.2}$ \\
& ChebNet  & 72.3$_{\pm 8.0}$  & 52.4$_{\pm 8.6}$  & 71.3$_{\pm 4.4}$  & 68.0$_{\pm 8.8}$  & - & \underline{77.6}$_{\pm 4.3}$  & 74.0$_{\pm 13.3}$ & 73.6$_{\pm 11.5}$ & \textbf{86.7}$_{\pm 0.8}$ \\
\bottomrule
\end{tabular}}
\caption{Generalization of different condensation methods across GNNs. (Result: average score $\pm$ standard deviation. \textbf{Bold}: best; \underline{Underline}: runner-up. -: cuda out of memory.)}
\label{table:cross-average}
\end{table*}

\subsection{Node Classification}
The node classification performance is reported in Table~\ref{table:performance_gcn}, where we reviewed the performance of the node classification task at different scales. For the needs of the multi-scale graph dataset condensation task, we distilled all graphs to the largest scale and then tested them by random sampling the subgraph according to the target reduction rate.

First, our model demonstrates strong performance across all scales. Notably, even with a subgraph containing just 10\% of the nodes from the largest condensed graph—a scenario that usually leads to a significant performance drop in other baseline models—our method maintains nearly lossless training performance. This suggests that our approach effectively mitigates challenges related to scale differences.

Second, our model not only maintains the training performance at different scales, but in some cases exceeds the performance of the original dataset. This remarkable result can be attributed to the information bottleneck refining process, in which we effectively retain the vast majority of low-frequency valid information while eliminating redundant data that may cause interference, which leads to improved generalization ability of the model.
% More experiment results are reported in the Appendix.

% Second, EXGC has better performance at different scales, but it can be observed that their effect at smaller scales, e.g., for each class of 1, is still more problematic, and there is a large gap with the training effect of the original dataset, which is in keeping with our theoretical analysis that the method limits the upper limit of the model.

\subsection{Cross-architecture Generalization}
Next, we evaluated the ability of the model's different scale condensation effects to generalize across models. For modeling, we used MLP~\cite{hu2021graphmlpnodeclassificationmessage}, SGC~\cite{wu2019simplifying}, APPNP~\cite{gasteiger2022predictpropagategraphneural} and ChebNet~\cite{he2024convolutionalneuralnetworksgraphs}. We tested the performance of the different compression models by sampling at four different size scales, and their means and variances are presented in the Table~\ref{table:cross-average}.
For their specific performance at each scale, we have reported in the Appendix. 
% For their specific performance at each scale, we have reported in the Appendix.
% A description of the model and its setup is in the Appendix.

First, our approach effectively eliminates scale differences across various model architectures, achieving the best average performance with the lowest variance, especially on large-scale graphs like Reddit. Even with drastic scale reductions, training performance remains consistent, proving the method's adaptability and universal applicability for multi-scale graph dataset condensation.

Second, when it comes to model generalization, our approach consistently achieves optimal results across various models. This indicates that, even while preserving the multi-scale effect, our method remains highly competitive at every scale. It is a good illustration of the advantages of our method in terms of generalization ability.
% \subsection{Time Comparison}
% Here, we comprehensively compare the scalability of our architecture to other compression models, as well as the time efficiency. Specifically, we transfer our framework to GCond and compare three other paradigms: recondensation, small-to-large, and large-to-small. The Ogbn-Arxiv dataset was used for the dataset. The results obtained are shown in Fig.
% First, our method demonstrates excellent performance on GCond, achieving results comparable to condensing each scale individually. While it is slightly less efficient than the one-time distillation method—due to the need to first process intermediate scales and then condense them in both directions—it remains significantly faster than re-condensation.
% \begin{figure*}[!t]
% \centering
% \subfigure[Ablation study about IB with different meso-scales.]{\label{fig:exp_ablation}
% \includegraphics[width=.45\linewidth]{figs/experiment_ablation.pdf}
% }
% \subfigure[Sensitivity study on meso-scale selection.]{\label{fig:exp_sensitivity}
% \includegraphics[width=.45\linewidth]{./figs/experiment_errorbar.pdf}
% }
% \centering
% \caption{Results of ablation and sensitivity study.}
% \label{fig:exp}
% \end{figure*}
% \subsection{Ablation and Sensitivity Study}

% \begin{figure}[t] % 'htbp' options for positioning
%     \centering
%     \includegraphics[width=0.9\linewidth]{./figs/experiment_errorbar.pdf}
%     \caption{Sensitivity study on meso-scale selection.}
%     \label{fig:exp_sensitivity}
% \end{figure}
\subsection{Ablation and Sensitivity Study}
Given that our model selects the initial meso-scale for multi-scale condensation process guided by information bottleneck(IB), we evaluated the necessity of IB and our model's sensitivity towards initial meso-scale selection. Here, we set the scales considered by meso-scale to 0.2,0.5,0.8. 

\begin{figure}[t] % 'htbp' options for positioning
    \centering
    \includegraphics[width=.8\linewidth]{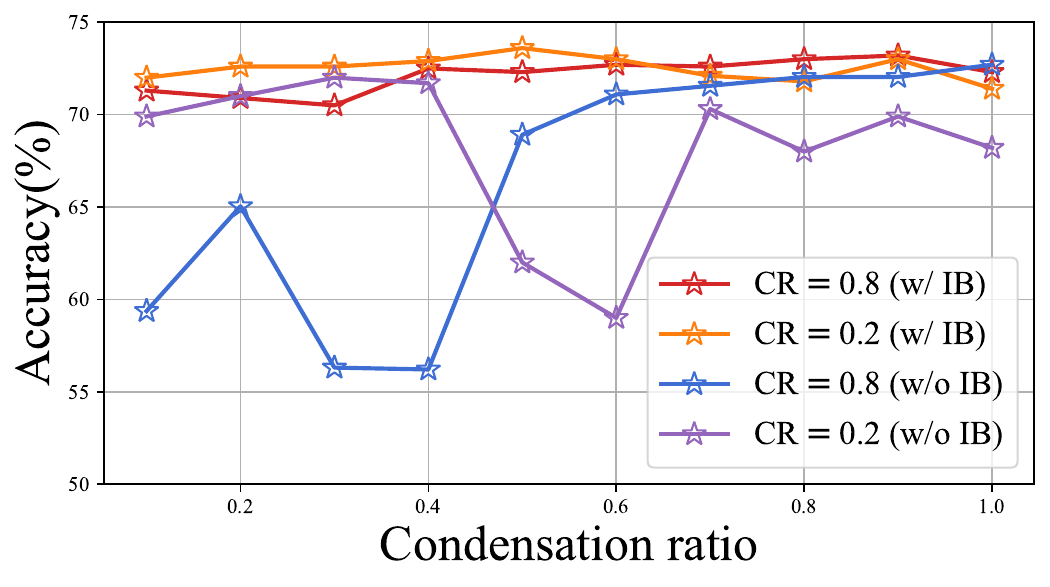}
    \caption{Ablation study for IB with different meso-scales.}
    \label{fig:exp_ablation}
\end{figure}

% Here we show the effect of different meso-scale settings on the effect.
% We conducted ablation experiments on the Citeseer dataset, setting the maximum scale to $r=1.0\%$ and adjusting various intermediate scales, both with and without the correlation loss of the information bottleneck. 
\textbf{Ablation study.}
For the ablation study on the impact of using IB, we set the maximum scale to $1.0\%$ and condensed the Citeseer dataset to various target reduction scales. Results for meso-scales of 0.2 and 0.8 are shown in Figure \ref{fig:exp_ablation}, indicating
that IB optimization effectively mitigates degradation caused by scale differences. Without IB, subgraph degradation occurs at a meso-scale of 0.8 with small compression rates, and larger scales fluctuate at a meso-scale of 0.2. With IB, performance across scales stabilizes.

\begin{figure}[t] % 'htbp' options for positioning
    \centering
    \includegraphics[width=0.8\linewidth]{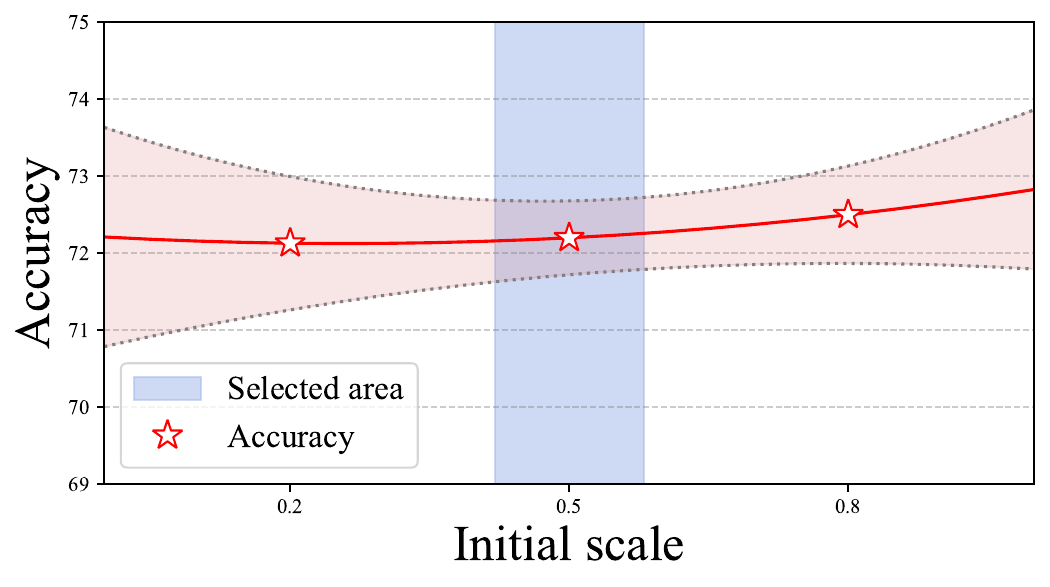}
    \caption{Sensitivity study on meso-scale selection.}
    \label{fig:exp_sensitivity}
\end{figure}

% Then during the generation of smaller graphs, useful information might be discarded, leading to a decline in performance.
% However, as the initial scale continues to rise, this effect might be hedged by the influence of much larger capacity which indicates complete original information, 
% Meanwhile, starting from a small initial scale might omit valuable information due to limit of graph size and preserve less original structure, causing the accuracy to drop.
% \begin{figure}[t] % 'htbp' options for positioning
%     \centering
%     \includegraphics[width=0.9\linewidth]{./figs/experiment_errorbar.pdf}
%     \caption{\bn{todo}}
%     \label{fig:exp_sensitivity}
% \end{figure}

\textbf{Meso-scale selection analysis.}
To evaluate how different initial scales affect the performance of multi-scale condensation, we calculated the average values and standard deviation based on the accuracy of node classification at all scales, forming Fig.\ref{fig:exp_sensitivity}.
The results indicate that as the initial meso-scale increases, the average performance varies by less than 0.5\%, with the lowest standard error observed at meso scales. This suggests that initial scale influences the consistency of bi-directional condensation, and our model maintains strong performance across all initial scales. We conclude that our IB-based adaptive selection of meso-scales ensures consistent performance across multiple scales with minimal average performance differences.

\section{Conclusion}

In this paper, we propose a GNN-centric Bi-directional Multi-scale Graph Dataset Condensation framework (\modelname) devised to achieve effective and efficient multi-scale graph condensation by unifying small-to-large and large-to-small paradigms. 
\modelname~starts with generating a meso-scale subgraph under the guidance of Informational Bottleneck principles, and then synthesize graphs at other scales by pruning or expanding based on the meso-scale subgraph.
% \modelname starts with generating a subgraph of a meso-scale which is estimated under the guidance of Informational Bottleneck, while graphs at other scales are synthesized by pruning or expanding based on the meso-scale subgraph.
Sufficient experiments were conducted to evaluate the performance of \modelname~in node classification and cross-architecture generalization tasks, with results demonstrating the clear superiority of \modelname.

\section*{Acknowledgments}
The corresponding author is Xingcheng Fu. Authors of this paper are supported by the National Natural Science Foundation of China through grants No.U21A20474 and No.62462007, and the Research Fund of Guangxi Key Lab of Multi-source Information Mining \& Security (24-A-02-01). We extend our sincere thanks to all authors for their valuable efforts and contributions.

\bibliography{reference}

\clearpage
\appendix
% \counterwithin{table}{section}
% % \counterwithin{footnote}{section}
% \counterwithin{figure}{section}
% \counterwithin{equation}{section}

% \section{Appendix}
\section{Derivation of Subgraph Condensation Information Bottleneck}

Here, we give a proof of the upper and lower bound derivation of the subgraph condensation information bottleneck. Its formulation can be expressed as:

\begin{equation}
\begin{aligned}
\mathop{\mathrm{max}}_{ G_{sub}\in {Sub(G)}}I\left(G_{sub};H(G,Y)\right)-\beta I\left(G;G_{sub}\right),
\end{aligned}
\label{ib2}
\end{equation}

Using the gradient-based method as an example, we replace $H(G,Y)$ with $\nabla_{\theta}$ for simplicity of representation. Thus, the objective can be reformulated as:

\begin{equation}
\begin{aligned}
\mathop{\mathrm{max}}_{ G_{sub}\in {Sub(G)}}I\left(G_{sub}\nabla_{\theta}\right)-\beta I\left(G;G_{sub}\right),
\end{aligned}
\label{gib}
\end{equation}

For the first part of the Eq~\eqref{gib}:

\begin{equation}
\begin{aligned}
I(G_{sub};\nabla_{\theta})&=\iint p(\nabla_{\theta},G_{sub})\log\frac{p(\nabla_{\theta},G_{sub})}{p(\nabla_{\theta})p(G_{sub})}\mathrm{d}\nabla_{\theta}\mathrm{d}G_{sub}
\\&=\iint p(\nabla_{\theta},G_{sub})\log\frac{p(\nabla_{\theta}|G_{sub})}{p(\nabla_{\theta})}\mathrm{d}\nabla_{\theta}\mathrm{d}G_{sub}
\\&=E\left[\log\frac{P\left(G_{sub}\mid\nabla_{\theta}^{\prime}\right)}{P\left(\nabla_{\theta}^{\prime}\right)}\right].
\end{aligned}
\end{equation}

However, since estimating $P\left(G_{sub}\mid\nabla_{\theta}^{\prime}\right)$ is challenging, it is common to use a variational approximation $Q\left(G_{sub}\mid\nabla_{\theta}^{\prime}\right)$ as a substitute.

\begin{equation}
\begin{aligned}
I\left(G_{sub};\nabla_{\theta}\right)& =E\left[\log\frac{P\left(\nabla_{\theta}\mid G_{sub}\right)}{P(\nabla_{\theta})}\right] 
\\&= E\left[\log\frac{P\left(\nabla_{\theta}\mid G_{sub}\right)Q\left(\nabla_{\theta}\mid G_{sub}\right)}{Q\left(\nabla_{\theta}\mid G_{sub}\right)P(\nabla_{\theta})}\right] 
\\&=E\left[\log\left(Q\left(\nabla_{\theta}|G_{sub}\right) 
\right)\right] \\
&\space+E[\mathrm{KL}\left(P\left(\nabla_{\theta}|G_{sub}\right)\|Q\left(\nabla_{\theta}|G_{sub}\right)\right)] \\
&\geq E\left[\log\left(Q\left(\nabla_{\theta}|G_{sub}\right)\right)\right].
\end{aligned}
\end{equation}

For the second part of the  Eq\eqref{gib}:

\begin{equation}
\begin{aligned}
I(G_{sub};G)&=\iint p(G,G_{sub})\log\frac{p(G,G_{sub})}{p(G)p(G_{sub})}\mathrm{d}G\mathrm{d}G_{sub}
\\&=\iint p(G,G_{sub})\log\frac{p(G_{sub}|G)}{p(G_{sub})}\mathrm{d}G\mathrm{d}G_{sub}
\\&=E\left[\log\frac{P\left(G\mid G_{sub}\right)}{P\left(G_{sub}\right)}\right].
\end{aligned}
\end{equation}

Similarly, since estimating $P(G_{sub})$ is difficult, we introduce an  variational approximation $R(G_{sub})$ as a substitute.

\begin{equation}
\begin{aligned}
I(G,G_{sub})&=E\left[\log\frac{P\left(G_{sub}\mid G\right)}{P\left(G_{sub}\right)}\right]
\\&=E\left[\log\frac{P\left(G_{sub}\mid G\right)}{R\left(G_{sub}\right)}\right]
\\&-\mathrm{KL}\left(P\left(G_{sub}\right)\parallel R\left(S_{sub}\right)\right)
\\&\leq E\left[\mathrm{KL}\left(P\left(G_{sub}\mid G\right)\parallel R\left(G_{sub}\right)\right)\right].
\end{aligned}
\end{equation}

\section{Experiment Details}
\subsection{Details of Datasets} 

We assess the performance of our model via experiments on a variety of datasets, including Citeseer, Cora, Ogbn-Arxiv, Flickr, Reddit. 

\begin{itemize}
    \item \textbf{Citation networks.} Citeseer, Cora~\cite{kipf2017semisupervised} and Ogbn-Arxiv~\cite{hu2021ogblsclargescalechallengemachine} are academic citation networks consisting of papers and citation relationships.
    \item \textbf{Social networks.} Flickr and Reddit~\cite{zeng2020graphsaintgraphsamplingbased} are social network datasets collected from social platforms Flickr and Reddit, in which nodes and edges stand for users and social interactions. 
\end{itemize}

Other information of the datasets is shown in Table \ref{table:datasets}.

\begin{table}[t]
\centering
\resizebox{\linewidth}{!}{
\begin{tabular}{lrrrr}
\toprule
\textbf{Dataset} & \textbf{\#Nodes} & \textbf{\#Edges} & \textbf{\#Classes} & \textbf{\#Features} \\
\midrule
Citeseer     & 3,327   & 4,732    & 6    & 3,703  \\
Cora         & 2,708   & 5,429    & 7    & 1,433  \\
Ogbn-Arxiv   & 169,343 & 1,166,243 & 40   & 128    \\
% \midrule
Flickr       & 89,250  & 899,756  & 7    & 500    \\
Reddit       & 232,965 & 57,307,946 & 210  & 602    \\
\bottomrule
\end{tabular}}
\caption{Statistics of Datasets}\label{table:datasets}
\end{table}

\begin{table*}[t] 
\centering
\resizebox{\textwidth}{!}{
\begin{tabular}{c|c|cccccccc|c}
\toprule
\multirow{2}{*}{Architecture} & \multirow{2}{*}{\begin{tabular}[c]{@{}c@{}}Reduction \\ rate(\%)\end{tabular}} & \multicolumn{9}{c}{Condensation Methods} \\ \cline{3-11} 
 & & \raisebox{-1mm}{\textbf{GCond}} & \raisebox{-1mm}{\textbf{SFGC}} & \raisebox{-1mm}{\textbf{SGDD}} & \raisebox{-1mm}{\textbf{EXGC}} & \raisebox{-1mm}{\textbf{GC-SNTK}} & \raisebox{-1mm}{\textbf{GCDM}} & \raisebox{-1mm}{\textbf{GDEM}} & \raisebox{-1mm}{\textbf{GEOM}} & \raisebox{-1mm}{\textbf{Ours}} \\
\midrule
\multirow{4}{*}{\textbf{GCN}} 
& 0.50 & 51.90$_{\pm 0.97}$ & 33.34$_{\pm 2.03}$ & 41.73$_{\pm 9.21}$ & 45.88$_{\pm 8.23}$ & 40.26$_{\pm 13.05}$& 51.80$_{\pm 2.55}$ & \underline{68.10}$_{\pm 2.54}$ & 32.28$_{\pm 5.93}$ & \textbf{73.70}$_{\pm 0.92}$\\
& 1.00 & 53.38$_{\pm 1.44}$ & 32.94$_{\pm 1.76}$ & 52.24$_{\pm 7.21}$ & 59.39$_{\pm 7.14}$ & 47.81$_{\pm 8.19}$ & 58.20$_{\pm 0.21}$ & \underline{70.94}$_{\pm 0.99}$ & 47.52$_{\pm 8.07}$ & \textbf{73.62}$_{\pm 0.63}$ \\
& 1.50 & 62.62$_{\pm 0.72}$ & 54.48$_{\pm 0.57}$ & 50.95$_{\pm 7.00}$ & 61.54$_{\pm 8.15}$ & 55.65$_{\pm 8.93}$ & 62.90$_{\pm 0.83}$ & \underline{71.78}$_{\pm 1.13}$ & 66.04$_{\pm 4.24}$ & \textbf{73.60}$_{\pm 1.85}$\\
& 2.00 & 69.80$_{\pm 0.25}$ & 60.16$_{\pm 0.60}$ & 70.90$_{\pm 0.31}$ & 71.04$_{\pm 0.64}$ & 65.37$_{\pm 5.07}$ & 65.70$_{\pm 1.10}$ & 72.66$_{\pm 0.30}$ & \textbf{74.26} $_{\pm 0.13}$ & \underline{73.60}$_{\pm 0.23}$\\
\midrule
\multirow{4}{*}{\textbf{MLP}}
& 0.50   & 48.66$_{\pm 3.32}$ & 23.34$_{\pm 2.03}$ & 50.28$_{\pm 7.25}$ & 58.17$_{\pm 3.64}$ & 35.70$_{\pm 8.52 }$&   32.06$_{\pm 8.67}$    & \underline{72.14}$_{\pm 0.48}$ & 32.28$_{\pm 5.93}$ & \textbf{73.62}$_{\pm 0.67}$ \\
& 1.00   & 62.84$_{\pm 0.69}$ & 22.88$_{\pm 1.74}$ & 55.23$_{\pm 5.66}$ & 59.45$_{\pm 3.80}$ & 50.75$_{\pm 12.97}$&   57.70$_{\pm 0.10}$    & \underline{72.98}$_{\pm 0.60}$ & 47.51$_{\pm 8.08}$ & \textbf{73.23}$_{\pm 0.53}$ \\
& 1.50   & 58.04$_{\pm 1.20}$ & 44.60$_{\pm 0.40}$ & 59.01$_{\pm 9.17}$ & 61.33$_{\pm 5.94}$ & 56.68$_{\pm 7.51 }$&   53.00$_{\pm 0.07}$    & \textbf{73.30}$_{\pm 0.47}$ & 66.02$_{\pm 4.23}$ & \underline{72.59}$_{\pm 0.97}$  \\
& 2.00   & 61.90$_{\pm 7.77}$ & 60.18$_{\pm 0.54}$ & 58.71$_{\pm 6.91}$ & 62.24$_{\pm 4.57}$ & 66.10$_{\pm 2.52 }$&   59.56$_{\pm 4.04}$    & 72.96$_{\pm 0.13}$ & \textbf{74.28}$_{\pm 0.13}$ & \underline{73.31}$_{\pm 0.81}$\\
\midrule
\multirow{4}{*}{\textbf{SGC}}
& 0.50   & 40.92$_{\pm 3.02}$ & 19.12$_{\pm 2.63}$ & 45.99$_{\pm 4.49}$ & 42.95$_{\pm 5.90}$ & 35.48$_{\pm 7.26 }$& 26.14$_{\pm 1.36}$ & \underline{72.10}$_{\pm 0.39}$ & 33.60$_{\pm 6.27}$ & \textbf{73.52}$_{\pm 0.26}$ \\
& 1.00   & 50.70$_{\pm 0.05}$ & 44.30$_{\pm 7.80}$ & 53.26$_{\pm 12.00}$ & 58.21$_{\pm 8.06}$ & 46.36$_{\pm 13.15}$ & 26.56$_{\pm 2.24}$ & \underline{72.34}$_{\pm 0.36}$ & 48.15$_{\pm 8.32}$ & \textbf{73.41}$_{\pm 0.41}$\\
& 1.50   & 55.22$_{\pm 1.23}$ & 48.58$_{\pm 6.84}$ & 50.76$_{\pm 5.31}$ & 65.20$_{\pm 4.93}$ & 63.21$_{\pm 3.55 }$& 27.32$_{\pm 2.95}$ & \underline{72.34}$_{\pm 0.52}$ & 65.81$_{\pm 4.50}$ & \textbf{73.24}$_{\pm 0.33}$ \\
& 2.00   & 55.72$_{\pm 3.85}$ & 48.42$_{\pm 9.78}$ & 69.94$_{\pm 0.33}$ & 70.84$_{\pm 0.36}$ & 67.81$_{\pm 2.77 }$& 32.60$_{\pm 2.77}$ & 72.30$_{\pm 0.57}$ & \textbf{74.22}$_{\pm 0.20}$ & \underline{73.47}$_{\pm 0.47}$\\
\midrule
\multirow{4}{*}{\textbf{APPNP}} 
& 0.50   & 30.72$_{\pm 1.73}$ & 25.08$_{\pm 5.49}$ & 33.10$_{\pm 9.80}$ & 32.30$_{\pm 9.62}$ & 22.82$_{\pm 11.91}$& 60.28$_{\pm 1.60}$ & \underline{70.96}$_{\pm 1.71}$ & 31.97$_{\pm 5.06}$ & \textbf{73.52}$_{\pm 2.31}$\\
& 1.00   & 41.64$_{\pm 1.03}$ & 30.38$_{\pm 4.56}$ & 56.88$_{\pm 7.90}$ & 55.98$_{\pm 7.55}$ & 32.42$_{\pm 5.13 }$& 60.08$_{\pm 3.49}$ & \underline{71.56}$_{\pm 0.47}$ & 34.87$_{\pm 5.93}$ & \textbf{73.35}$_{\pm 0.24}$\\
& 1.50   & 63.50$_{\pm 0.40}$ & 37.12$_{\pm 6.97}$ & 64.20$_{\pm 4.50}$ & 63.38$_{\pm 4.25}$ & 45.86$_{\pm 2.66 }$& 61.76$_{\pm 2.36}$ & \underline{71.60}$_{\pm 0.59}$ & 42.02$_{\pm 8.37}$ & \textbf{72.91}$_{\pm 2.31}$ \\
& 2.00   & \underline{72.04}$_{\pm 0.86}$ & 48.48$_{\pm 0.38}$ & 70.60$_{\pm 0.20}$ & 69.66$_{\pm 0.08}$ & 54.02$_{\pm 0.48 }$& 68.08$_{\pm 1.46}$ & 71.74$_{\pm 0.15}$ & 65.68$_{\pm 0.24}$ & \textbf{73.40}$_{\pm 0.32}$ \\
\midrule
\multirow{4}{*}{\textbf{ChebNet}} 
& 0.50   & 35.62$_{\pm 6.75}$ & 34.06$_{\pm 6.88}$ & 40.60$_{\pm 3.89}$ & 59.92$_{\pm 3.71}$ & 46.10$_{\pm 11.48}$&   38.70$_{\pm 3.50}$    & \underline{71.10}$_{\pm 1.52}$ & 38.56$_{\pm 8.95}$ & \textbf{72.91}$_{\pm 1.62}$\\
& 1.00   & 50.84$_{\pm 5.50}$ & 40.88$_{\pm 0.68}$ & 53.40$_{\pm 6.14}$ & 62.70$_{\pm 5.98}$ & 59.26$_{\pm 7.20 }$&   43.70$_{\pm 2.95}$    & \underline{71.68}$_{\pm 1.06}$ & 52.29$_{\pm 6.63}$ & \textbf{73.24}$_{\pm 0.92}$\\
& 1.50   & 55.54$_{\pm 5.78}$ & 57.04$_{\pm 1.22}$ & 63.40$_{\pm 4.38}$ & 65.70$_{\pm 4.27}$ & 67.62$_{\pm 2.26 }$&  36.82$_{\pm 7.60}$     & \underline{72.00}$_{\pm 0.97}$ & 66.01$_{\pm 4.10}$ & \textbf{73.18}$_{\pm 0.84}$\\
& 2.00   & 47.22$_{\pm 2.51}$ & 67.10$_{\pm 0.34}$ & 64.50$_{\pm 1.00}$ & 68.84$_{\pm 0.87}$ & 69.12$_{\pm 1.30 }$&   30.34$_{\pm 3.45}$    & 72.90$_{\pm 0.11}$ & \textbf{74.09}$_{\pm 0.13}$ & \underline{73.40}$_{\pm 0.53}$\\
\bottomrule
\end{tabular}}   
\caption{Performance comparison across different backbones and reduction rates on CiteSeer dataset. (Result: average score $\pm$ standard deviation. \textbf{Bold}: best; \underline{Underline}: runner-up.)}
\label{table:cross_architecture_citeseer}
\end{table*}

\subsection{Details of Backbones} 
Also, we choose a variety of GNNs, i.e., GCN\cite{kipf2017semisupervised} , MLP~\cite{hu2021graphmlpnodeclassificationmessage}, SGC \cite{wu2019simplifying} , APPNP~\cite{gasteiger2022predictpropagategraphneural}  and ChebNet~\cite{he2024convolutionalneuralnetworksgraphs} for cross-model experiments.

\begin{itemize}
    \item MLP~\cite{hu2021graphmlpnodeclassificationmessage} is a basic neural network backbone consisting of fully connected layers, commonly employed as a baseline in graph tasks where structural information is not leveraged. 
    \item SGC~\cite{wu2019simplifying} is a streamlined version of Graph Convolutional Networks (GCNs) that reduces complexity by removing non-linearities and collapsing weight matrices, making it more efficient for certain graph tasks. 
    \item APPNP~\cite{gasteiger2022predictpropagategraphneural} integrates personalized PageRank with neural networks to effectively propagate labels across a graph, enhancing node classification performance by accounting for both global and local structures. 
    \item ChebNet~\cite{he2024convolutionalneuralnetworksgraphs} is a spectral graph convolutional model that utilizes Chebyshev polynomials to efficiently approximate filters in the frequency domain, enabling localized and scalable graph learning.
\end{itemize}

\subsection{Details of Baselines} 
For evaluation, we compare BiMSGC with state-of-the-art graph learning methods:
\begin{itemize}
    \item GCond \cite{jin2022graph} optimizes a gradient matching loss to imitate the training trajectory on the original graph and condensing node features and structural information.
    \item SFGC \cite{zheng2023structurefree} distills a small-scale graph node set of structure-free data by encoding the topology structure information into the node attributes.
    \item SGDD \cite{yang2023does} preserves the original graph structure by broadcasting it during the generation of synthetic condensed graphs. 
    \item EXGC \cite{fang2024exgc} employs Mean-Field variational approximation and integrates explanation techniques to prune redundancy and improve compactness.
    \item GC-SNTK \cite{wang2024fast} reformulates condensation as a Kernel Ridge Regression task, capturing the topology with a structure-based neural tangent kernel.
    \item GCDM \cite{liu2022graphcondensationreceptivefield} optimizes synthetic graphs to match the distribution of receptive fields in the original graph based on an informative representation.
    \item GEOM \cite{zhang2024navigating} trains expert trajectories with supervision signals and utilizes an expanding window matching strategy to transfer information.
    \item GDEM \cite{liu2024graph} minimizes the spectral domain differences by aligning the eigenbasis and node features of real and synthetic graphs.
\end{itemize}

\subsection{Details of Implementation} 
All experiments were conducted with a two-layer GCN as the training target, where the hidden layer dimension was uniformly set to 256. The first step distills the meso-scale with epoch set to 500s, distills to the small scale with epoch set to 50s, and distills to the large scale with epoch set to 500s.
For the rest of the hyperparameters we align with the settings provided in GDEM\cite{liu2024graph}. The testing epoch was set to 2000.
We conducted 5 repeated trials for each setting, and calculated the mean value and standard deviation. 
We set the learning rate of our model to 0.0001, and run 2000 epochs for testing. 
All models were trained and tested on a single Nvidia A800 80GB GPU.

\section{Cross-Model Experiment}
\label{app:cross}

\begin{table*}[t]
\centering
\resizebox{\textwidth}{!}{
\begin{tabular}{c|c|cccccccc|c}
\toprule
\multirow{2}{*}{Architecture} & \multirow{2}{*}{\begin{tabular}[c]{@{}c@{}}Reduction \\ rate(\%)\end{tabular}} & \multicolumn{9}{c}{Condensation Methods} \\ \cline{3-11} 
 & & \raisebox{-1mm}{\textbf{GCond}} & \raisebox{-1mm}{\textbf{SFGC}} & \raisebox{-1mm}{\textbf{SGDD}} & \raisebox{-1mm}{\textbf{EXGC}} & \raisebox{-1mm}{\textbf{GC-SNTK}} & \raisebox{-1mm}{\textbf{GCDM}} & \raisebox{-1mm}{\textbf{GDEM}} & \raisebox{-1mm}{\textbf{GEOM}} & \raisebox{-1mm}{\textbf{Ours}} \\
\midrule
\multirow{4}{*}{\textbf{GCN}}
& 0.50 & 58.00$_{\pm 5.41}$ & 66.92$_{\pm 7.01}$ & 61.60$_{\pm 6.31}$ & 61.47$_{\pm 7.20}$ & 47.39$_{\pm 3.34}$ & 53.23$_{\pm 2.68}$ & \underline{68.68}$_{\pm 3.62}$ & 39.92$_{\pm 10.41}$ &\textbf{78.22}$_{\pm 1.25}$\\
& 1.00 & 73.40$_{\pm 4.63}$ & 70.32$_{\pm 0.64}$ & \underline{74.36}$_{\pm 2.03}$ & 71.89$_{\pm 2.45}$ & 65.39$_{\pm 10.01}$ & 61.25$_{\pm 4.84}$ & 69.94$_{\pm 1.67}$& 65.97$_{\pm 6.55}$ & \textbf{80.46}$_{\pm 0.96}$\\
& 1.50 & 79.16$_{\pm 1.24}$ & \underline{79.30}$_{\pm 0.51}$ & 75.49$_{\pm 2.27}$ & 77.93$_{\pm 1.22}$ & 71.65$_{\pm 4.87}$ & 55.97$_{\pm 2.97}$ & 71.08$_{\pm 2.88}$ & 77.62$_{\pm 2.12}$ & \textbf{79.47}$_{\pm 1.44}$ \\
& 2.00 & 80.04$_{\pm 0.27}$ & 80.82$_{\pm 0.43}$ & 77.91$_{\pm 1.00}$ & 79.81$_{\pm 0.38}$ & 74.85$_{\pm 2.59}$ & 78.98$_{\pm 3.01}$ & 74.08$_{\pm 2.90}$ & \textbf{83.64}$_{\pm 0.18}$ & \underline{80.91}$_{\pm 0.62}$ \\
\midrule
\multirow{4}{*}{\textbf{MLP}}
& 0.50   & 67.10$_{\pm 6.15}$ & \underline{70.8}4$_{\pm  8.64}$ & 61.31$_{\pm 6.76}$ & 63.49$_{\pm  4.92}$ & 42.50$_{\pm 13.94}$ &     42.36$_{\pm 1.35}$        & 59.52$_{\pm 4.61}$& 39.90$_{\pm 10.43}$ & \textbf{79.43}$_{\pm 2.25}$\\
& 1.00   & 71.98$_{\pm 1.90}$ & \underline{78.24}$_{\pm  1.40}$ & 72.47$_{\pm 3.00}$ & 70.83$_{\pm  2.91}$ & 66.36$_{\pm  8.96}$ &     55.84$_{\pm 2.25}$        & 67.88$_{\pm 1.66}$ & 65.97$_{\pm 6.49}$ & \textbf{79.81}$_{\pm 1.31}$\\
& 1.50   & 72.02$_{\pm 4.38}$ & \textbf{80.22}$_{\pm  1.20}$ & 75.77$_{\pm 2.67}$ & 73.56$_{\pm  3.19}$ & 70.25$_{\pm  4.72}$ &     48.98$_{\pm 1.06}$        & 71.88$_{\pm 2.47}$ & 77.71$_{\pm 2.13}$ & \underline{78.32}$_{\pm 3.49}$\\
& 2.00   & 71.32$_{\pm 3.37}$ & \underline{80.82}$_{\pm  0.43}$ & 78.22$_{\pm 0.49}$ & 76.47$_{\pm  2.48}$ & 74.40$_{\pm  2.24}$ &      63.32$_{\pm 0.95}$       & 75.70$_{\pm 1.08}$ & \textbf{83.65}$_{\pm 0.18}$ & 78.95$_{\pm 0.75}$\\
\midrule
\multirow{4}{*}{\textbf{SGC}} 
& 0.50   & 42.38$_{\pm 2.39}$ & 71.54$_{\pm 2.14}$ & 57.70$_{\pm 8.07}$ & 55.77$_{\pm 10.39}$ & 55.64$_{\pm 11.58}$ & 24.20$_{\pm6.55}$  & \underline{75.10}$_{\pm 1.30}$ & 39.65$_{\pm 10.42}$& \textbf{78.52}$_{\pm 1.14}$\\
& 1.00   & 63.96$_{\pm 1.53}$ & \underline{77.44}$_{\pm 1.80}$ & 73.07$_{\pm 4.76}$ & 72.46$_{\pm  3.31}$ & 64.55$_{\pm  8.72}$ & 34.84$_{\pm2.01}$  & 74.94$_{\pm 1.70}$ & 65.54$_{\pm 6.05 }$& \textbf{80.71}$_{\pm 1.02}$\\
& 1.50   & 75.72$_{\pm 0.10}$ & \underline{78.64}$_{\pm 0.97}$ & 75.20$_{\pm 2.53}$ & 77.11$_{\pm  2.21}$ & 69.55$_{\pm  5.38}$ & 21.20$_{\pm1.23}$  & 73.96$_{\pm 2.05}$ & 77.72$_{\pm 2.15 }$& \textbf{79.70}$_{\pm 2.25}$\\
& 2.00   & 77.48$_{\pm 1.21}$ & 78.84$_{\pm 0.46}$ & 78.17$_{\pm 0.82}$ & 78.81$_{\pm  0.51}$ & 71.88$_{\pm  2.70}$ & 54.80$_{\pm0.00}$  & 73.92$_{\pm 1.29}$ & \textbf{83.14}$_{\pm 0.43 }$& \underline{79.70}$_{\pm 0.28}$ \\
\midrule
\multirow{4}{*}{\textbf{APPNP}} 
& 0.50   & 46.60$_{\pm 12.13}$ & \underline{72.74}$_{\pm 4.07}$ & 58.30$_{\pm 7.40}$ & 57.54$_{\pm  7.07}$ & 29.09$_{\pm 10.33}$ &   50.80$_{\pm 8.15}$ & 50.68$_{\pm 7.67}$ & 34.98$_{\pm 8.81 }$& \textbf{79.81}$_{\pm 4.32}$\\
& 1.00   & 53.00$_{\pm 1.35}$ & 76.14$_{\pm  1.54}$ & 62.68$_{\pm 3.00}$ & 71.94$_{\pm  2.83}$ & 29.86$_{\pm 13.85}$ &  \underline{77.74}$_{\pm 0.73}$   & 68.76$_{\pm 3.85}$& 38.07$_{\pm 9.11 }$& \textbf{80.43}$_{\pm 2.18}$ \\
& 1.50   & 74.14$_{\pm 0.84}$ & \underline{78.48}$_{\pm  1.20}$ & 77.90$_{\pm 2.30}$ & 77.12$_{\pm  2.16}$ & 52.55$_{\pm 16.40}$ & 72.66$_{\pm 1.87}$   & 70.60$_{\pm 2.63}$ & 41.85$_{\pm 8.85 }$& \textbf{79.02}$_{\pm 1.52}$\\
& 2.00   & 77.00$_{\pm 1.18}$ & \textbf{81.12}$_{\pm  0.24}$ & \underline{80.48}$_{\pm 0.18}$ & 79.50$_{\pm  0.06}$ & 46.94$_{\pm  0.67}$ &  70.90$_{\pm 0.78}$  & 74.40$_{\pm 0.45}$ & 67.33$_{\pm 0.21 }$& 80.23$_{\pm 0.46}$\\
\midrule
\multirow{4}{*}{\textbf{ChebNet}} 
& 0.50   & 60.96$_{\pm 0.92}$ & \underline{70.40}$_{\pm  1.77}$ & 65.10$_{\pm 4.15}$ & 64.36$_{\pm  4.08}$ & 54.10$_{\pm  6.81}$ &   30.38$_{\pm 8.95}$          & 55.24$_{\pm 4.80}$ & 46.40$_{\pm 7.82}$ & \textbf{74.73}$_{\pm 1.83}$\\
& 1.00   & 65.16$_{\pm 1.92}$ & 74.48$_{\pm  0.92}$ & \underline{74.50}$_{\pm 1.40}$ & 73.72$_{\pm  1.22}$ & 69.19$_{\pm  4.79}$ &   47.28$_{\pm 10.22}$         & 68.70$_{\pm 3.26}$ & 65.03$_{\pm 6.88}$ & \textbf{74.72}$_{\pm 2.45}$\\
& 1.50   & 73.54$_{\pm 0.90}$ & \textbf{78.06}$_{\pm  0.91}$ & 75.90$_{\pm 1.30}$ & 76.14$_{\pm  1.11}$ & 73.88$_{\pm  3.56}$ &    40.64$_{\pm 7.77}$         & 71.66$_{\pm 2.14}$ & 75.77$_{\pm 1.92}$ & \underline{76.41}$_{\pm 0.62}$ \\
& 2.00   & 76.66$_{\pm 0.82}$ & \underline{79.80}$_{\pm  0.51}$ & 76.30$_{\pm 0.85}$ & 77.50$_{\pm  0.70}$ & 73.44$_{\pm  1.96}$ &    40.88$_{\pm 9.00}$         & 76.38$_{\pm 0.71}$ & \textbf{80.73}$_{\pm 0.80}$ & 75.20$_{\pm 0.38}$ \\
\bottomrule
\end{tabular}}
\caption{Performance comparison across different backbones and reduction rates on Cora dataset. (Result: average score $\pm$ standard deviation. \textbf{Bold}: best; \underline{Underline}: runner-up.)}
\label{table:cross_architecture_cora}
\end{table*}

% \caption{Node classification performance of different condensation methods,. (Result: average score $\pm$ standard deviation. \textbf{Bold}: best; \underline{Underline}: runner-up.)}
% \caption{Generalization of different condensation methods across GNNs. (Result: average score $\pm$ standard deviation. \textbf{Bold}: best; \underline{Underline}: runner-up.)}

\begin{table*}[t]
\centering
\resizebox{\textwidth}{!}{
\begin{tabular}{c|c|cccccccc|c}
\toprule
\multirow{2}{*}{Architecture} & \multirow{2}{*}{\begin{tabular}[c]{@{}c@{}}Reduction \\ rate(\%)\end{tabular}} & \multicolumn{9}{c}{Condensation Methods} \\ \cline{3-11} 
 & & \raisebox{-1mm}{\textbf{GCond}} & \raisebox{-1mm}{\textbf{SFGC}} & \raisebox{-1mm}{\textbf{SGDD}} & \raisebox{-1mm}{\textbf{EXGC}} & \raisebox{-1mm}{\textbf{GC-SNTK}} & \raisebox{-1mm}{\textbf{GCDM}} & \raisebox{-1mm}{\textbf{GDEM}} & \raisebox{-1mm}{\textbf{GEOM}} & \raisebox{-1mm}{\textbf{Ours}} \\
\midrule
\multirow{4}{*}{\textbf{GCN}} 
& 0.05 & 48.80$_{\pm 3.98}$ & 45.31$_{\pm 3.05}$ & 52.18$_{\pm 0.92}$ & 44.11$_{\pm 3.96}$ & 25.14$_{\pm 5.61}$ &  51.89$_{\pm 1.70}$ & \underline{61.02}$_{\pm 1.96}$ & 44.86$_{\pm 4.41}$ & \textbf{62.68}$_{\pm 0.42}$ \\
& 0.10 & 54.01$_{\pm 0.72}$ & 52.63$_{\pm 0.76}$ & 60.15$_{\pm 0.34}$ & 51.03$_{\pm 3.07}$ & 25.14$_{\pm 5.61}$ &  60.31$_{\pm 2.90}$ &\underline{60.49}$_{\pm 2.21}$  & 50.55$_{\pm 3.34}$ & \textbf{63.20}$_{\pm 0.21}$\\
& 0.30 & 60.15$_{\pm 1.15}$ & 62.33$_{\pm 1.22}$ & \underline{63.43}$_{\pm 0.09}$ & 59.18$_{\pm 1.96}$ & 46.09$_{\pm 3.83}$ &  62.41$_{\pm 2.74}$ & 60.65$_{\pm 1.41}$ & \textbf{63.75}$_{\pm 0.91}$ & 62.72$_{\pm 0.15}$\\
& 0.50 & 64.53$_{\pm 0.09}$ & \underline{66.24}$_{\pm 0.41}$ & 63.42$_{\pm 0.34}$ & 62.78$_{\pm 0.50}$ & 55.44$_{\pm 1.17}$ &  64.16$_{\pm 2.90}$ & 62.14$_{\pm 1.08}$ & \textbf{68.84}$_{\pm 0.17}$ & 63.69$_{\pm 0.18}$\\
\midrule
\multirow{4}{*}{\textbf{MLP}} 
& 0.05   & 48.38$_{\pm  3.07}$ & 47.45$_{\pm  3.96}$ & 54.27$_{\pm  0.35}$ & 44.02$_{\pm  3.00}$ & 25.17$_{\pm  5.56}$ & 49.35$_{\pm 0.48}$      & \underline{56.06}$_{\pm  0.68}$ & 44.87$_{\pm 4.38}$ & \textbf{61.32}$_{\pm  0.31}$ \\
& 0.10   & 51.34$_{\pm  2.11}$ & 53.47$_{\pm  2.32}$ & \underline{61.93}$_{\pm  0.12}$ & 50.35$_{\pm  3.73}$ & 30.17$_{\pm  5.56}$ & 46.77$_{\pm 1.90}$      & 58.46$_{\pm  1.38}$ & 50.49$_{\pm 3.30}$ & \textbf{63.35}$_{\pm  1.15}$ \\
& 0.30   & 58.81$_{\pm  1.81}$ & 62.64$_{\pm  1.71}$ & \underline{63.45}$_{\pm  0.33}$ & 59.56$_{\pm  1.36}$ & 43.83$_{\pm  4.48}$ & 55.61$_{\pm 0.40}$      & 59.39$_{\pm  0.63}$ & \textbf{63.77}$_{\pm 0.94}$ & 62.31$_{\pm  0.37}$\\
& 0.50   & 60.80$_{\pm  0.55}$ & \underline{66.03}$_{\pm  0.14}$ & 62.06$_{\pm  0.56}$ & 62.10$_{\pm  0.32}$ & 52.52$_{\pm  1.66}$ & 49.26$_{\pm 0.49}$      & 59.49$_{\pm  1.15}$ & \textbf{68.80}$_{\pm 0.16}$ & 62.48$_{\pm  0.33}$\\
\midrule
\multirow{4}{*}{\textbf{SGC}} 
& 0.05   & 44.57$_{\pm  3.83}$ & 45.73$_{\pm  3.54}$ & \underline{58.24}$_{\pm  0.56}$ & 42.33$_{\pm  3.08}$ & 25.71$_{\pm  7.26}$ & 50.86$_{\pm 0.14}$ & \textbf{63.23}$_{\pm  0.40}$ & 41.68$_{\pm 6.24}$ &  57.91$_{\pm  4.28}$\\
& 0.10   & 53.48$_{\pm  2.81}$ & 51.27$_{\pm  3.17}$ & \underline{61.89}$_{\pm  0.23}$ & 47.60$_{\pm  3.20}$ & 25.71$_{\pm  7.26}$ & 57.18$_{\pm 4.39}$ & \textbf{63.13}$_{\pm  0.35}$ & 48.29$_{\pm 1.46}$ &  58.72$_{\pm  0.23}$\\
& 0.30   & 60.29$_{\pm  2.19}$ & 61.69$_{\pm  1.16}$ & \underline{62.23}$_{\pm  0.19}$ & 58.80$_{\pm  1.60}$ & 51.66$_{\pm  4.13}$ & 61.90$_{\pm 0.96}$ & \textbf{63.20}$_{\pm  0.35}$ & 60.27$_{\pm 2.28}$ &  62.15$_{\pm  0.14}$\\
& 0.50   & 64.97$_{\pm  0.29}$ & \underline{65.82}$_{\pm  0.14}$ & 64.32$_{\pm  0.43}$ & 64.06$_{\pm  0.38}$ & 57.31$_{\pm  2.23}$ & 62.45$_{\pm 0.23}$ & 63.02$_{\pm  0.39}$ & \textbf{66.47}$_{\pm 0.15}$ &  62.15$_{\pm  0.17}$\\
\midrule
\multirow{4}{*}{\textbf{APPNP}} 
& 0.05   & 48.44$_{\pm  1.16}$ & 46.60$_{\pm  2.96}$ & 57.86$_{\pm  2.10}$ & 46.86$_{\pm  1.96}$ & 12.47$_{\pm  3.77}$ & 49.85$_{\pm 0.88}$ & \underline{57.91}$_{\pm  0.04}$ & 40.89$_{\pm 4.61}$ & \textbf{60.11}$_{\pm  0.03}$\\
& 0.10   & 54.16$_{\pm  1.34}$ & 50.77$_{\pm  2.95}$ & 56.20$_{\pm  2.60}$ & 51.20$_{\pm  2.41}$ & 31.69$_{\pm  4.90}$ & 56.54$_{\pm 0.30}$ & \underline{61.09}$_{\pm  0.33}$ & 48.34$_{\pm 2.31}$ & \textbf{63.93}$_{\pm  0.16}$ \\
& 0.30   & \textbf{62.40}$_{\pm  0.74}$ & 59.76$_{\pm  0.92}$ & 60.58$_{\pm  1.90}$ & 59.68$_{\pm  1.70}$ & 47.98$_{\pm  2.05}$ & 55.70$_{\pm 0.46}$ & 61.67$_{\pm  0.33}$ & 59.16$_{\pm 0.84}$ & \underline{61.82}$_{\pm  0.42}$\\
& 0.50   & \textbf{64.31}$_{\pm  0.13}$ & 62.22$_{\pm  0.29}$ & 63.90$_{\pm  0.40}$ & 63.05$_{\pm  0.27}$ & 51.93$_{\pm  0.01}$ & 55.36$_{\pm 1.77}$ & 61.17$_{\pm  0.02}$ & \underline{64.03}$_{\pm 0.09}$ & 61.15$_{\pm  0.04}$ \\
\midrule
\multirow{4}{*}{\textbf{ChebNet}} 
& 0.05   & 41.06$_{\pm  4.03}$ & 42.31$_{\pm  1.04}$ & 48.00$_{\pm  3.87}$ & 37.18$_{\pm  3.70}$ & 13.12$_{\pm  4.09}$ &   36.00$_{\pm 1.18}$    & \underline{55.82}$_{\pm  0.72}$ & 38.00$_{\pm 2.80}$ & \textbf{56.84}$_{\pm  0.26}$ \\
& 0.10   & 47.12$_{\pm  2.23}$ & 48.64$_{\pm  1.47}$ & 46.30$_{\pm  1.98}$ & 45.57$_{\pm  1.81}$ & 12.90$_{\pm  2.87}$ &   36.00$_{\pm 1.18}$    & \textbf{58.02}$_{\pm  0.60}$ & 44.92$_{\pm 0.66}$ & \underline{57.29}$_{\pm  0.17}$\\
& 0.30   & 56.68$_{\pm  0.91}$ & \underline{57.33}$_{\pm  1.19}$ & 54.20$_{\pm  1.00}$ & 53.41$_{\pm  0.84}$ & 29.93$_{\pm  3.86}$ &   42.24$_{\pm 0.99}$    & \textbf{57.99}$_{\pm  0.71}$ & 56.56$_{\pm 1.15}$ & 56.62$_{\pm  0.68}$\\
& 0.50   & 58.47$_{\pm  0.37}$ & \underline{61.45}$_{\pm  0.27}$ & 58.20$_{\pm  0.60}$ & 57.45$_{\pm  0.42}$ & 35.51$_{\pm  4.77}$ &   42.34$_{\pm 0.81}$    & 57.81$_{\pm  0.52}$ & \textbf{62.28}$_{\pm 0.17}$ & 56.73$_{\pm  0.09}$\\
\bottomrule
\end{tabular}}

\caption{Performance comparison across different backbones and reduction rates on Ogbn-Arxiv dataset. (Result: average score $\pm$ standard deviation. \textbf{Bold}: best; \underline{Underline}: runner-up.)}
\label{table:cross_architecture_arxiv}
\end{table*}

\begin{table*}[t]
\centering
\resizebox{\textwidth}{!}{
\begin{tabular}{c|c|cccccccc|c}
\toprule
\multirow{2}{*}{Architecture} & \multirow{2}{*}{\begin{tabular}[c]{@{}c@{}}Reduction \\ rate(\%)\end{tabular}} & \multicolumn{9}{c}{Condensation Methods} \\ \cline{3-11} 
 & & \raisebox{-1mm}{\textbf{GCond}} & \raisebox{-1mm}{\textbf{SFGC}} & \raisebox{-1mm}{\textbf{SGDD}} & \raisebox{-1mm}{\textbf{EXGC}} & \raisebox{-1mm}{\textbf{GC-SNTK}} & \raisebox{-1mm}{\textbf{GCDM}} & \raisebox{-1mm}{\textbf{GDEM}} & \raisebox{-1mm}{\textbf{GEOM}} & \raisebox{-1mm}{\textbf{Ours}} \\
\midrule
\multirow{4}{*}{\textbf{GCN}} 
& 0.10 & \underline{46.22}$_{\pm 2.10}$ & 44.09$_{\pm 1.53}$ & 43.99$_{\pm 2.07}$ & 42.23$_{\pm 1.27}$ & 28.35$_{\pm 5.71}$ & 32.95$_{\pm 2.47}$  & 42.61$_{\pm 2.45}$ & 43.03$_{\pm 1.71}$ & \textbf{50.34}$_{\pm 0.45}$\\
& 0.30 & \underline{46.80}$_{\pm 0.16}$ & 43.69$_{\pm 0.69}$ & 45.55$_{\pm 0.24}$ & 44.63$_{\pm 0.95}$ & 29.11$_{\pm 3.91}$ & 36.18$_{\pm 0.79}$ & 45.37$_{\pm 2.60}$  & 45.63$_{\pm 0.53}$ & \textbf{50.37}$_{\pm 0.03}$ \\
& 0.70 & 47.08$_{\pm 0.21}$ & 44.28$_{\pm 0.31}$ & \underline{47.97}$_{\pm 0.27}$ & 46.34$_{\pm 0.47}$ & 32.23$_{\pm 6.02}$ & 36.15$_{\pm 1.35}$ & 46.27$_{\pm 2.46}$  & 46.73$_{\pm 0.23}$ & \textbf{50.55}$_{\pm 0.32}$ \\
& 1.00 & 47.09$_{\pm 0.10}$ & 44.22$_{\pm 0.16}$ & 47.94$_{\pm 0.17}$ & 46.91$_{\pm 0.17}$ & 31.16$_{\pm 3.26}$ & 35.92$_{\pm 0.50}$  & \underline{49.43}$_{\pm 3.46}$ & 47.29$_{\pm 0.08}$ & \textbf{50.63}$_{\pm 0.13}$ \\
\midrule
\multirow{4}{*}{\textbf{MLP}} 
& 0.10   & 42.82$_{\pm  0.25}$ & 40.25$_{\pm  1.01}$ & 41.69$_{\pm  0.32}$ & 40.13$_{\pm  1.90}$ & 22.94$_{\pm  2.02}$ &   41.92$_{\pm 0.92}$    & 39.58$_{\pm  1.01}$ & \underline{43.03}$_{\pm 1.70}$ & \textbf{49.94}$_{\pm  0.87}$\\
& 0.30   & 43.48$_{\pm  0.96}$ & 41.89$_{\pm  0.84}$ & 42.42$_{\pm  1.04}$ & 42.04$_{\pm  1.26}$ & 23.24$_{\pm  3.95}$ &   41.32$_{\pm 1.79}$    & 39.41$_{\pm  1.32}$ & \underline{45.60}$_{\pm 0.51}$ & \textbf{49.04}$_{\pm  2.26}$\\
& 0.70   & 43.23$_{\pm  0.36}$ & 40.72$_{\pm  0.22}$ & 44.17$_{\pm  0.42}$ & 42.88$_{\pm  0.90}$ & 25.25$_{\pm  3.66}$ &   43.80$_{\pm 0.35}$    & 38.49$_{\pm  0.32}$ & \underline{46.73}$_{\pm 0.23}$ & \textbf{50.04}$_{\pm  0.78}$\\
& 1.00   & 42.17$_{\pm  0.33}$ & 41.00$_{\pm  0.84}$ & 44.03$_{\pm  0.39}$ & 42.52$_{\pm  0.69}$ & 30.25$_{\pm  3.86}$ &   44.26$_{\pm 0.61}$    & 40.09$_{\pm  1.54}$ & \underline{47.29}$_{\pm 0.09}$ & \textbf{49.75}$_{\pm  0.64}$ \\
\midrule
\multirow{4}{*}{\textbf{SGC}}  
& 0.10& \underline{46.26}$_{\pm  0.21}$ & 43.49$_{\pm  1.08}$ & 42.07$_{\pm  0.28}$ & 42.73$_{\pm  0.74}$ & 28.68$_{\pm  6.69}$ & 34.90$_{\pm 0.88}$ & 39.90$_{\pm 3.28}$ & 42.81$_{\pm 1.49}$ & \textbf{49.17}$_{\pm  2.94}$\\
& 0.30& 46.48$_{\pm  0.22}$ & 45.69$_{\pm  0.54}$ & \underline{47.35}$_{\pm  0.29}$ & 44.51$_{\pm  0.83}$ & 30.13$_{\pm  8.25}$ & 34.45$_{\pm 0.40}$ & 40.08$_{\pm  3.21}$ & 45.56$_{\pm 0.62}$ & \textbf{49.06}$_{\pm  0.83}$\\
& 0.70& \underline{46.83}$_{\pm  0.24}$ & 45.45$_{\pm  0.42}$ & 45.65$_{\pm  0.31}$ & 45.83$_{\pm  0.30}$ & 33.35$_{\pm  6.49}$ & 32.81$_{\pm 1.29}$ & 39.60$_{\pm  3.88}$ & 46.21$_{\pm 0.31}$ & \textbf{49.85}$_{\pm  3.28}$\\
& 1.00& \underline{46.93}$_{\pm  0.28}$ & 45.15$_{\pm  0.06}$ & 46.75$_{\pm  0.34}$ & 46.55$_{\pm  0.34}$ & 37.58$_{\pm  3.92}$ & 34.40$_{\pm 0.69}$ & 39.62$_{\pm  4.19}$ & 46.54$_{\pm 0.32}$ & \textbf{49.57}$_{\pm  1.92}$\\
\midrule
\multirow{4}{*}{\textbf{APPNP}}  
& 0.10& 26.95$_{\pm  1.93}$ & 32.72$_{\pm  2.40}$ & 38.60$_{\pm  2.20}$ & 27.83$_{\pm  2.05}$ & 17.85$_{\pm  1.43}$ &  \underline{41.41}$_{\pm 1.63}$ & 31.60$_{\pm  0.36}$ & 32.24$_{\pm 1.99}$ & \textbf{49.90}$_{\pm  0.62}$ \\
& 0.30& 28.85$_{\pm  1.77}$ & 35.98$_{\pm  1.11}$ & 40.65$_{\pm  1.50}$ & 29.65$_{\pm  1.34}$ & 19.68$_{\pm  1.04}$ &  \underline{41.85}$_{\pm 1.39}$ & 31.41$_{\pm  0.55}$ & 36.28$_{\pm 1.14}$ & \textbf{49.98}$_{\pm  1.24}$ \\
& 0.70& 32.75$_{\pm  0.98}$ & 37.35$_{\pm  0.75}$ & \underline{44.40}$_{\pm  0.90}$ & 33.60$_{\pm  0.84}$ & 23.99$_{\pm  0.65}$ &  43.79$_{\pm 0.29}$ & 31.69$_{\pm  0.29}$ & 38.89$_{\pm 0.59}$ & \textbf{49.92}$_{\pm  0.37}$ \\
& 1.00& 35.74$_{\pm  0.22}$ & 37.97$_{\pm  0.05}$ & 43.49$_{\pm  0.30}$ & 35.49$_{\pm  0.27}$ & 24.97$_{\pm  0.17}$ &  \underline{44.31}$_{\pm 0.68}$ & 31.34$_{\pm  0.32}$ & 39.95$_{\pm 0.29}$ & \textbf{49.89}$_{\pm  0.09}$ \\
\midrule
\multirow{4}{*}{\textbf{ChebNet}}  
& 0.10& 39.84$_{\pm  2.72}$ & 40.15$_{\pm  1.63}$ & 40.20$_{\pm  2.00}$ & 39.36$_{\pm  1.83}$ & 24.32$_{\pm  4.67}$ &  \underline{40.30}$_{\pm 1.68}$ & 38.96$_{\pm  2.06}$ & 38.26$_{\pm 2.24}$ & \textbf{45.33}$_{\pm  2.03}$ \\
& 0.30& 41.54$_{\pm  0.79}$ & 40.50$_{\pm  1.11}$ & 41.60$_{\pm  1.35}$ & 40.76$_{\pm  1.28}$ & 27.48$_{\pm  4.95}$ &  41.52$_{\pm 1.46}$ & 38.06$_{\pm  1.07}$ & \underline{42.36}$_{\pm 1.71}$ & \textbf{45.61}$_{\pm  0.73}$ \\
& 0.70& 43.25$_{\pm  0.68}$ & 41.41$_{\pm  1.47}$ & 42.50$_{\pm  1.30}$ & 41.61$_{\pm  1.12}$ & 28.50$_{\pm  4.79}$ &  41.97$_{\pm 1.81}$ & 38.48$_{\pm  2.35}$ & \underline{44.46}$_{\pm 0.42}$ & \textbf{46.52}$_{\pm  2.26}$ \\
& 1.00& 43.67$_{\pm  0.24}$ & 41.03$_{\pm  1.03}$ & 42.20$_{\pm  0.50}$ & 41.34$_{\pm  0.38}$ & 28.05$_{\pm  3.71}$ &  42.67$_{\pm 1.08}$ & 37.46$_{\pm  1.30}$ & \underline{44.91}$_{\pm 0.13}$ & \textbf{46.43}$_{\pm  2.68}$ \\
\bottomrule
\end{tabular}}

\caption{Performance comparison across different backbones and reduction rates on Flickr dataset. (Result: average score $\pm$ standard deviation. \textbf{Bold}: best; \underline{Underline}: runner-up.)}
\label{table:cross_architecture_flickr}
\end{table*}

\begin{table*}[t]
\centering
\resizebox{\textwidth}{!}{
\begin{tabular}{c|c|ccccccc|c}
\toprule
\multirow{2}{*}{Architecture} & \multirow{2}{*}{\begin{tabular}[c]{@{}c@{}}Reduction \\ rate(\%)\end{tabular}} & \multicolumn{8}{c}{Condensation Methods} \\ \cline{3-10} 
 & & \raisebox{-1mm}{\textbf{GCond}} & \raisebox{-1mm}{\textbf{SFGC}} & \raisebox{-1mm}{\textbf{SGDD}} & \raisebox{-1mm}{\textbf{EXGC}} 
 % & \raisebox{-1mm}{\textbf{GC-SNTK}} 
 & \raisebox{-1mm}{\textbf{GCDM}} & \raisebox{-1mm}{\textbf{GDEM}} & \raisebox{-1mm}{\textbf{GEOM}} & \raisebox{-1mm}{\textbf{Ours}} \\
\midrule
\multirow{4}{*}{\textbf{GCN}}
& 0.05 & 73.02$_{\pm 5.27}$ & 64.44$_{\pm 0.83}$ & 80.97$_{\pm 0.94}$ & 71.46$_{\pm 2.99}$ &  77.95$_{\pm 3.46}$ & \underline{83.57}$_{\pm 0.97}$ & 68.98$_{\pm 5.73}$ & \textbf{93.29}$_{\pm 0.73}$ \\

& 0.10 & 84.17$_{\pm 1.50}$ & 67.57$_{\pm 0.25}$ & 82.75$_{\pm 0.20}$ & 82.69$_{\pm 2.12}$ & \underline{88.86}$_{\pm 1.69}$  & 86.60$_{\pm 1.00}$ & 83.92$_{\pm 1.28}$ & \textbf{93.61}$_{\pm 0.21}$\\

& 0.15 & 87.40$_{\pm 0.79}$ & 80.26$_{\pm 0.32}$ & 81.94$_{\pm 0.75}$ & 87.09$_{\pm 1.01}$ & 85.87$_{\pm 0.39}$ & 86.47$_{\pm 0.35}$  & \underline{88.92}$_{\pm 0.67}$ & \textbf{92.91}$_{\pm 0.15}$\\

& 0.20 & 90.91$_{\pm 0.23}$ & 84.58$_{\pm 1.91}$ & 88.47$_{\pm 0.23}$ & 89.37$_{\pm 0.31}$ & 89.23$_{\pm 0.52}$  & \textbf{93.14}$_{\pm 0.14}$ & 91.48$_{\pm 0.08}$ & \underline{93.05}$_{\pm 0.11}$ \\
\midrule
\multirow{4}{*}{\textbf{MLP}} 
& 0.05   & 42.18$_{\pm  0.36}$ & 47.45$_{\pm  3.96}$ & 42.98$_{\pm  0.42}$ & 37.16$_{\pm  1.90}$ &  66.94$_{\pm 0.21}$     & 43.62$_{\pm  0.25}$ & \underline{69.00}$_{\pm 5.74}$ & \textbf{93.01}$_{\pm  0.47}$\\

& 0.10   & 46.37$_{\pm  0.38}$ & 53.47$_{\pm  2.32}$ & 47.12$_{\pm  0.45}$ & 42.52$_{\pm  1.42}$ &  76.53$_{\pm 0.42}$     & 57.77$_{\pm  0.62}$ & \underline{83.94}$_{\pm 1.28}$ & \textbf{93.27}$_{\pm  0.59}$ \\

& 0.15   & 52.84$_{\pm  1.61}$ & 62.64$_{\pm  1.71}$ & 52.68$_{\pm  1.68}$ & 44.92$_{\pm  1.16}$ &  79.15$_{\pm 8.80}$    & 58.34$_{\pm  0.36}$ & \underline{88.94}$_{\pm 0.69}$ & \textbf{93.26}$_{\pm  1.41}$ \\

& 0.20   & 56.61$_{\pm  1.36}$ & 66.03$_{\pm  0.14}$ & 58.49$_{\pm  1.42}$ & 46.61$_{\pm  0.75}$ &  88.18$_{\pm 6.64}$   & 59.36$_{\pm  0.46}$ & \underline{91.48}$_{\pm 0.07}$ & \textbf{93.11}$_{\pm  0.78}$\\

\midrule
\multirow{4}{*}{\textbf{SGC}}
& 0.05   & 77.68$_{\pm  0.68}$ & 45.73$_{\pm  3.54}$ & \underline{85.76}$_{\pm  0.38}$ & 71.30$_{\pm  4.54}$ & 82.98$_{\pm 0.04}$ & 83.48$_{\pm  0.65}$ & 67.35$_{\pm 6.68}$ & \textbf{88.75}$_{\pm  0.72}$\\

& 0.10   & \underline{88.36}$_{\pm  0.56}$ & 51.27$_{\pm  3.17}$ & 86.85$_{\pm  0.42}$ & 85.01$_{\pm  2.43}$ & 79.97$_{\pm 0.69}$ & 86.16$_{\pm  0.38}$ & 83.76$_{\pm 2.05}$ & \textbf{89.02}$_{\pm  0.15}$\\

& 0.15   & \underline{89.13}$_{\pm  0.59}$ & 61.69$_{\pm  1.16}$ & 80.23$_{\pm  0.60}$ & 87.75$_{\pm  1.25}$ & 85.26$_{\pm 5.74}$ & 86.53$_{\pm  0.65}$ & 88.07$_{\pm 0.86}$ & \textbf{89.95}$_{\pm  0.48}$ \\

& 0.20   & \textbf{91.13}$_{\pm  0.12}$ & 65.82$_{\pm  0.14}$ & 83.15$_{\pm  1.84}$ & 89.22$_{\pm  0.43}$ & 87.18$_{\pm 4.39}$ & 89.65$_{\pm  0.46}$ & \underline{90.57}$_{\pm 0.19}$ & 90.01$_{\pm  1.62}$ \\
\midrule
\multirow{4}{*}{\textbf{APPNP}} 
& 0.05   & 72.17$_{\pm  1.38}$ & 55.01$_{\pm  0.89}$ & 72.80$_{\pm  1.90}$ & 71.97$_{\pm  1.58}$ &  \underline{74.54}$_{\pm 2.68}$     & 56.83$_{\pm  0.01}$ & 68.90$_{\pm 4.03}$ & \textbf{92.02}$_{\pm  0.24}$\\

& 0.10   & 82.55$_{\pm  1.03}$ & 57.29$_{\pm  0.17}$ & \underline{86.68}$_{\pm  2.00}$ & 81.96$_{\pm  1.75}$ &   83.96$_{\pm 2.69}$    & 82.93$_{\pm  0.03}$ & 81.32$_{\pm 1.24}$ & \textbf{92.56}$_{\pm  0.17}$ \\

& 0.15   & 85.39$_{\pm  0.52}$ & 69.69$_{\pm  0.14}$ & 82.56$_{\pm  1.30}$ & 85.56$_{\pm  1.22}$ &   81.24$_{\pm 5.72}$    & 82.91$_{\pm  0.07}$ & \underline{86.79}$_{\pm 0.31}$ & \textbf{92.51}$_{\pm  2.15}$ \\

& 0.20   & 87.59$_{\pm  0.01}$ & 82.87$_{\pm  0.61}$ & 89.08$_{\pm  0.20}$ & 88.16$_{\pm  0.03}$ &   \underline{90.82}$_{\pm 2.74}$    & 88.07$_{\pm  0.01}$ & 89.87$_{\pm 0.07}$ & \textbf{92.16}$_{\pm  0.24}$\\

\midrule
\multirow{4}{*}{\textbf{ChebNet}}
& 0.05   & 61.07$_{\pm  2.28}$ & 42.31$_{\pm  1.04}$ & 66.30$_{\pm  4.68}$ & 55.50$_{\pm  4.59}$ &   \underline{76.21}$_{\pm 0.62}$    & 54.15$_{\pm  1.25}$ & 57.19$_{\pm 5.48}$ &\textbf{86.71}$_{\pm  2.35}$ \\

& 0.10   & 72.66$_{\pm  0.79}$ & 48.64$_{\pm  1.47}$ & 69.40$_{\pm  1.70}$ & 68.60$_{\pm  1.58}$ &   \underline{83.78}$_{\pm 7.72}$    & 77.74$_{\pm  3.25}$ & 74.37$_{\pm 1.23}$ &\textbf{85.50}$_{\pm  2.73}$ \\

& 0.15   & 75.81$_{\pm  0.77}$ & 57.33$_{\pm  1.19}$ & 72.90$_{\pm  1.15}$ & 72.11$_{\pm  0.98}$ &   73.80$_{\pm 6.39}$    & \underline{81.82}$_{\pm  0.34}$ & 79.31$_{\pm 1.09}$ &\textbf{87.44}$_{\pm  0.68}$ \\

& 0.20   & 79.78$_{\pm  0.41}$ & 61.45$_{\pm  0.27}$ & 76.60$_{\pm  1.15}$ & 75.83$_{\pm  0.96}$ &  76.40$_{\pm 6.21}$     & 82.31$_{\pm  2.50}$ & \underline{83.36}$_{\pm 0.34}$ &\textbf{87.12}$_{\pm  1.04}$  \\
\bottomrule
\end{tabular}}
\caption{Performance comparison across different backbones and reduction rates on Reddit dataset. (Result: average score $\pm$ standard deviation. \textbf{Bold}: best; \underline{Underline}: runner-up. GC-SNTK omitted for cuda out of memory.)}
\label{table:cross_architecture_reddit}
\end{table*}

\subsubsection{Model Setting.}

For SGC, MLP and APPNP, we employ 2-layer aggregations, with the dimension of hidden layer set to 256. For ChebNet, we configure the convolution layers to $2$ with propagation steps selected from $\{2, 3, 5\}$. 
Considering the difference across datasets, we first distilled them to the largest reduction-rate scale and then sample from them on demand for testing. Specifically,
we set the reduction rate for Citeseer and Cora to be of $\{0.5\%,1.0\%,1.5\%,2.0\%\}$, while the reduction rate for Ogbn-Arxiv is $\{0.05\%,0.1\%,0.3\%,0.5\%\}$. The reduction rate of Flickr and Reddit is respectively 
$\{0.1\%,0.3\%,0.7\%,1\%\}$ and 
$\{0.05\%,0.1\%,0.15\%,0.2\%\}$.
For each setting, we also conducted 5 repeated trials and computed the mean value and standard deviation.

\subsubsection{Performance.}
% f>r>ci>co>o
Across various datasets, our approach consistently achieves optimal results. 
% First, our approach effectively eliminates scale differences across various model architectures, achieving the best average performance with the lowest variance, especially on large-scale graphs like Reddit. Even with significant scale reductions, training performance remains consistent, demonstrating the method’s adaptability and universal applicability for multi-scale graph dataset condensation.
% Second, in terms of model generalization, our approach consistently delivers optimal results across various models. This indicates that our method remains highly competitive at every scale while preserving the multi-scale effect. It exemplifies the advantages of our method in terms of generalization ability.
On large datasets with high-dimensional of node features such as Flickr and Reddit, our approach exhibits superior performance, as shown in Table \ref{table:cross_architecture_flickr} and \ref{table:cross_architecture_reddit}. In comparison, on smaller and sparser graphs like Citeseer and Cora (shown in Table \ref{table:cross_architecture_citeseer} and \ref{table:cross_architecture_cora}), while our method still outperforms others, the stability of performance has slightly declined, which might be attributed to the original valuable signal being weaker and thus harder to capture fully. Finally, though our model maintain highest average rank on Ogbn-Arxiv (Table \ref{table:cross_architecture_arxiv}), it did not always achieve top performance for every reduction rate and backbone, potentially due to the low dimensionality of node features.

\end{document}